\documentclass[journal]{IEEEtran}
\usepackage[utf8]{inputenc}
\usepackage[hmargin=1in,vmargin=1in,bindingoffset=0.0cm]{geometry}
\usepackage{times}
\usepackage{amsmath,amsthm,amsfonts,amssymb,fixmath,dsfont,bm}
\usepackage{graphicx}
\usepackage{xcolor}
\usepackage{xspace}
\usepackage{mathtools}
\usepackage{cite}
\usepackage{multirow}
\usepackage{bigstrut}
\usepackage{graphicx}
\usepackage{epstopdf}
\usepackage{algpseudocode}
\usepackage{algorithm}
\usepackage{authblk}
\usepackage[hyphens]{url}
\usepackage{enumitem}
\usepackage{mdframed}
\usepackage{tabularx}

\usepackage{tikz}
\tikzstyle{guide}=[]
\usetikzlibrary{arrows,positioning}

\newcommand{\ie}{i.e.,\xspace}

\renewcommand{\paragraph}[1]{\vspace{3mm}\noindent\textbf{#1}}

\renewcommand{\b}[1]{{\mathbold{#1}}}   
\renewcommand{\(}{\left(}           
\renewcommand{\)}{\right)}

\newcommand{\x}{\b{x}}
\newcommand{\xhat}{\hat{\b{x}}}

\renewcommand{\u}{\b{u}}

\renewcommand{\L}{\b{L}}            
\newcommand{\U}{\b{U}}              
\newcommand{\bLambda}{\b{\Lambda}}  
\newcommand{\bOmega}{\b{\Omega}}    
\newcommand{\M}{\b{M}}
\newcommand{\A}{\b{A}}

\newcommand{\I}{\b{I}}
\newcommand{\X}{\b{X}}
\newcommand{\Xhat}{\widehat{\X}}
\newcommand{\C}{\b{C}}
\newcommand{\Y}{\b{Y}}
\newcommand{\Z}{\b{Z}}
\newcommand{\D}{\b{D}}              

\newcommand{\W}{\b{W}}              
\newcommand{\K}{\b{K}}

\newcommand{\Vs}{\mathcal{V}}
\newcommand{\Es}{\mathcal{E}}

\newcommand{\Dy}[1]{{D}_{#1}}

\newcommand{\tl}{\theta_{\ell}}%
\renewcommand{\ll}{\lambda_{\ell}}%
\newcommand{\lmax}{\lambda_{\mathrm{max}}}%
\newcommand{\kk}{\omega_{k}}%
\newcommand{\eu}{e}
\newcommand{\ju}{j}

\newcommand*{\Scale}[2][4]{\scalebox{#1}{#2}}%

\newcommand{\LG}{\L_{\hspace{-1px}G}}              
\newcommand{\LJ}{\L_{\hspace{-1px}J}}              
\newcommand{\LT}{\L_{\hspace{-1px}T}}              

\newcommand{\TG}[2]{{\mathcal{T}_{#2}^{\hspace{1px}G}} #1 \hspace{0.2mm}}              
\newcommand{\TJ}[2]{{\mathcal{T}_{#2}^{\hspace{1px}J}} \hspace{0.5mm} #1 \hspace{0.2mm}}              

\newcommand{\Ker}[2]{\K_{#1}{\hspace{-.5mm}\lrbrace{#2}}} 
\newcommand{\GKer}[2]{\widetilde{\K}_{#1}{\hspace{-.5mm}\lrbrace{#2}}} 

\newcommand{\UG}{\U_{\hspace{-1px}G}}              
\newcommand{\UJ}{\U_{\hspace{-1px}J}}              
\newcommand{\UT}{\U_{\hspace{-1px}T}}              

\newcommand{\GFT}[1]{\mathrm{GFT}\hspace{-.0mm}\{#1\}}
\newcommand{\IGFT}[1]{\mathrm{GFT}^{\hspace{0.2mm}\Scale[0.7]{-1}}\hspace{-.0mm}\{#1\}}
\newcommand{\DFT}[1]{\mathrm{DFT}\hspace{-.0mm}\{#1\}}

\newcommand{\JFT}[1]{\mathrm{JFT}\hspace{-.0mm}\{#1\}}
\newcommand{\IJFT}[1]{\mathrm{JFT}^{\hspace{0.2mm}\Scale[0.7]{-1}}\hspace{-0.0mm}\{#1\}}
\newcommand{\STVFT}[1]{\mathrm{STVFT}\hspace{-0.0mm}\{#1\}}
\newcommand{\STVWT}[1]{\mathrm{STVWT}\hspace{-0.0mm}\{#1\}}

\newcommand{\jnorm}[3]{\mathrm{N}_{#2,#3}\lrpar{#1}}

\newcommand{\Rbb}{\mathbb{R}} 
\newcommand{\Cbb}{\mathbb{C}}

\renewcommand{\vec}[1]{\textrm{vec}\hspace{-.5mm}\(#1\)}           
\newcommand{\mat}[1]{\textrm{mat}\hspace{-.5mm}\(#1\)}           
\newcommand{\diag}[1]{\textrm{diag}\hspace{-.5mm}\(#1\)}           
\DeclareMathOperator*{\argmin}{arg\,min}

\DeclarePairedDelimiter{\norm}{\lVert}{\rVert}
\DeclarePairedDelimiter\abs{\lvert}{\rvert}%

\makeatletter
\newcommand{\parder}[2]{\begingroup
  \@tempswafalse\toks@={}\count@=\z@
  \@for\next:=#2\do
    {\expandafter\check@var\next\@nil
     \advance\count@\parder@exp
     \if@tempswa
       \toks@=\expandafter{\the\toks@\,}%
     \else
       \@tempswatrue
     \fi
     \toks@=\expandafter{\the\expandafter\toks@\expandafter\partial\parder@var}}%
  \frac{\partial\ifnum\count@=\@ne\else^{\number\count@}\fi#1}{\the\toks@}%
  \endgroup}
\def\check@var{\@ifstar{\mult@var}{\one@var}}
\def\mult@var#1#2\@nil{\def\parder@var{#2^{#1}}\def\parder@exp{#1}}
\def\one@var#1\@nil{\def\parder@var{#1}\chardef\parder@exp\@ne}
\makeatother

\newcommand{\grad}{\nabla}
\newcommand{\GG}{\grad_{\hspace{-1px}G}}              
\newcommand{\GJ}{\grad_{\hspace{-1px}J}}              
\newcommand{\GT}{\grad_{\hspace{-1px}T}}              
\newcommand{\divG}{\GG^{\hermitian}}

\newcommand{\trace}[1]{\textrm{tr}\hspace{-.5mm}\(#1\)}            
\newcommand{\transpose}{\intercal}                      
\newcommand{\hermitian}{*}                      
\newcommand{\delequal}{\overset{\Delta}{=}} 

\newtheorem{theorem}{Theorem}

\newcommand{\lrpar}[1]{\left( {#1} \right)}
\newcommand{\lrsquare}[1]{\left[ {#1} \right]}
\newcommand{\lrbrace}[1]{\left\lbrace{#1}\right\rbrace}

\graphicspath{ {image/} }

\author[1]{Francesco Grassi}
\author[2]{Andreas Loukas}
\author[2]{Nathana\"{e}l Perraudin}
\author[2]{Benjamin Ricaud\vspace*{-1.5ex}}
\affil[1]{Politecnico di Torino, Torino, Italy}
\affil[2]{Ecole Polytechnique F\'ed\'erale de Lausanne, Lausanne, Switzerland}

\title{A Time-Vertex Signal Processing Framework\\[0.2em] \large{Scalable processing and meaningful representations for time-series on graphs}} 

\begin{document}
\maketitle
\begin{abstract}
An emerging way to deal with high-dimensional non-euclidean data is to assume that the underlying structure can be captured by a graph. Recently, ideas have begun to emerge related to the analysis of time-varying graph signals. 
This work aims to elevate the notion of joint harmonic analysis to a full-fledged framework denoted as Time-Vertex Signal Processing, that links together the time-domain signal processing techniques with the new tools of graph signal processing. 
This entails three main contributions: 
(a) We provide a formal motivation for harmonic time-vertex analysis as an analysis tool for the state evolution of simple Partial Differential Equations on graphs. 
(b) We improve the accuracy of joint filtering operators by up-to two orders of magnitude. 
(c) Using our joint filters, we construct time-vertex dictionaries analyzing the different scales and the local time-frequency content of a signal.    
The utility of our tools is illustrated in numerous applications and datasets, such as dynamic mesh denoising and classification, still-video inpainting, and source localization in seismic events. Our results suggest that joint analysis of time-vertex signals can bring benefits to regression and learning.
\end{abstract}
\begin{IEEEkeywords}
Time-Vertex Signal Processing, Graph Signal Processing, Partial Differential Equations
\end{IEEEkeywords}

\section{Introduction}

Whether examining consensus and rumor spreading over social networks~\cite{de2013anatomy,adamic2005political,guille2013information}, transportation networks~\cite{mohan2008nericell} and related epidemic spreading~\cite{RevModPhys.87.925}, neuronal activation patterns~\cite{huang2015graph} or time-evolving functional network in the brain \cite{smith2016physiological}, as well as other datasets collected from a variety of fields, such as physics, engineering, and life-science, much of the high-dimensional data exhibit complex non-euclidean properties. 

An emerging way to deal with these issues is to use a graph to capture the structure underlying the data. This has been the driving force behind recent efforts in the signal processing field to extend harmonic analysis to graph signals, \ie signals supported on the vertices of irregular graphs~\cite{shuman2013emerging, sandryhaila2013discrete}.  
In the field of graph signal processing (GSP), the introduction of the graph Fourier transform (GFT) has enabled us to perform harmonic analysis taking into account the structure of the data, and has lead to improvements for tasks such as clustering~\cite{belkin2001laplacian},
low-rank extraction~\cite{shahid2016fast}, spectral estimation~\cite{perraudin2016stationary,marques2016stationary}, non-stationary analysis~\cite{hammond2011wavelets,coifman2006diffusion} and semi-supervised learning~\cite{belkin2004semi,smola2003kernels}.

Nevertheless, though state-of-the-art graph-based methods have been successful for many tasks, so far they predominantly ignore the time-dimension of data, for example by treating successive signals independently or performing a global average~\cite{huang2015graph,perraudin2016stationary,kalofolias2016learn}. On the contrary, many of the systems to which GSP is applied to are inherently dynamic.
Consider for instance a road network, and suppose that we want to infer traffic conditions given flow information over a subset of highways and streets. Approaches that do not take into account the temporal evolution of traffic will be biased by seasonal variations and unable to provide insights about transient phenomena, such as rush hour traffic, bottlenecks caused by blockage, and stop waves.

Recently, several ideas begin to emerge related to the analysis of time-varying graph signals, such as Joint time-vertex Fourier transform (JFT)~\cite{loukas2016frequency} and the joint time-vertex filters~\cite{isufi2016separable} and filterbanks~\cite{grassi2016tracking}. While these constitute notable contributions, we argue that the potential of joint harmonic analysis is yet unexplored, both in terms of its {foundations}, {algorithms}, and {applications}.

In this work we aim at elevating the notion of joint harmonic analysis to a full-fledged framework, referred to as the \emph{Time-Vertex Signal Processing Framework}, that links together the time-domain signal processing techniques with the new tools of GSP.  

\vspace{2mm} \noindent
This entails the following contributions:

\noindent 1. \emph{Connection to PDEs.} 
We illustrate how joint analysis emerges when analyzing the state evolution of simple Partial Differential Equations (PDEs) on graphs (Section~\ref{ssec:linearPDE}). We also provide an example (epidemic spreading) demonstrating that the joint frequency analysis can be meaningful for the study non-linear and stochastic processes leading to a compact and intuitive representation (Section~\ref{ssec:epidem}).

\noindent 2. \emph{Accurate fast joint filtering.} In Section~\ref{sec:Jop} we illustrate the utility of joint filtering time-vertex signals and propose a fast filtering implementation, called Fast Fourier-Chebyshev (FFC) algorithm, which significantly improves upon the state-of-the-art filters both in terms of separable and non-separable filtering objectives. For the latter case especially, our numerical experiments show that FFC can yield up to two orders of magnitude smaller approximation error at a similar complexity to previous joint filters.

\noindent 3. \emph{Overcomplete representations.} We study redundant time-vertex dictionaries and exploit them for signal analysis and synthesis. The proposed framework includes a frame condition guaranteeing that no information is lost. Two particular cases are: \textit{time-vertex wavelets} capturing the different scales of the signal components, and the \textit{short time-vertex Fourier transform} that is useful in determining the local time frequency content of the signal. 

\vspace{2mm}\noindent\emph{4. Illustrating the utility of time-vertex analysis.} 
Finally, Section~\ref{sec:eval} provides experimental evidence for the utility of joint harmonic analysis in a number of graph-temporal datasets that were up to now not fully exploited, such as dynamic meshes, video and general dynamics over networks.  
The range of applications covers the classical signal processing problems of denoising, inpainting and compression, but also extends to feature extraction for classification and source localization problems. 
%

\subsection{Related Work}

The time-vertex framework is intimately linked with the stochastic analysis of multivariate signals and, therefore, with graphical models (e.g. \cite{dahlhaus2003causality} and references therein).
The main difference between graphical models and GSP lies in the assumption about the relation between the signal and graph~\cite{zhang2015graph}. Graphical models adopt a purely Bayesian setting, where edges denote conditional dependencies between variables. As such, the graph usually is a proxy for the covariance and is learned from the data. On the other hand, GSP assumes that the graph is given and its relation to the signal can be understood through harmonic analysis. 

In this context, the idea of time-vertex analysis can be traced back to the study~\cite{sandryhaila2014big} aiming to process multi-modal signals with different graphs associated with each of their modalities (i.e., one can consider a time-vertex signal as multi-modal, with time and graph being the two modalities). 
Collaboration between the graph theory and signal processing communities has led to new tools to process and analyze time-varying graphs and signals on a graph, such as multilayer graphs and tensor products of graphs~\cite{kivela2014multilayer,benzi2016principal,de2013mathematical}.
The notion of joint time-vertex harmonic analysis was further realized in~\cite{loukas2016frequency} by one of the authors of this work. Therein, the joint Fourier analysis is presented and its properties analyzed in details, together with examples of joint filters. In this work, we leverage these concepts proposing a framework in which the joint Fourier transform is just one of the building blocks.

\paragraph{Visualization, filtering and stationarity.} 
The idea of analyzing the behavior of graph filters with time-varying signals first appeared in~\cite{loukas2015distributed}, showing that graph filters could be analyzed by applying jointly a GFT and a Z-transform and as such they possess a joint frequency response. Since then, we have seen a number of works dealing with time-varying signals on graphs: 
Authors in~\cite{valdivia2015wavelet} propose a method that relies on graph wavelet theory and product graphs to visualize time-varying data defined on the vertices of a graph in order to identify spatial and/or temporal variations.
A step towards the graphical model has been carried out by authors in~\cite{mei2015signal}. In this work, authors assume data time dependencies to be modeled by an auto-regressive (AR) process and they propose several algorithms to estimate the network structure capturing the spatio-temporal dependencies and the coefficients of the AR process expressed as graph polynomial filters.
In order to deal with the high computational complexity of the eigendecomposition, different filtering approximation algorithms have been proposed, mainly based on polynomials: centralized and distributed joint filter 2D Chebychev polynomial~\cite{loukas2016frequency}, separable rational~\cite{isufi2016separable} implementations, and autoregressive models~\cite{isufi2017autoregressive}.

Finally, in parallel with this work, the authors extended the notions of time stationarity and the recent graph stationarity~\cite{perraudin2016stationary} to the joint time-vertex domain~\cite{perraudin2016towards} providing a framework for the statistical signal processing of time-vertex signals. Authors showed that assuming joint stationarity to regularize learning can yield significant accuracy improvements and reduce computational complexity in both estimation and recovery tasks prediction with respect to purely time or graph methods~\cite{loukas2016predicting,loukas2016stationary}.
Despite the relevance of this work to time-vertex analysis, here we focus on the purely deterministic setting. 

\section{Harmonic Time-Vertex Analysis}\label{sec:TVR}

We denote by $G=(\Vs,\Es,\W_G)$ the graph, where $\Vs$ indicates the set of nodes, $\Es$ the set of edges and $\W_G$ is the associated $N\times N$ weight matrix. Furthermore, let ${\LG = \D_G -\W_G}$ be the  combinatorial Laplacian matrix, i.e. the finite difference approximation to the continuous Laplacian operator \cite{smola2003kernels} or the Laplace-Beltrami operator for Riemannian manifolds \cite{burago2013graph}.
We suppose that the signal on a graph is sampled at $T$ successive regular intervals of unit length. That is, if we denote by $\x_t \in \Rbb^N$ the graph signal at instant $t$, the time-varying graph signal corresponds to the matrix $\X = \left[ \x_1, \x_2, \ldots, \x_T \right] \in \Rbb^{N\times T}$. 
We denote $\X^\transpose$, $\bar{\X}$, and $\X^\hermitian$ the transpose, the complex conjugate and the hermitian of $\X$. Furthermore, we refer to both $\X$ and its vectorized form $\x = \vec{\X} \in \Rbb^{NT}$ as ``{time-vertex signal}''.

\subsection{The joint time-vertex Fourier transform}

The main idea of harmonic analysis is to decompose a signal into oscillating modes thanks to the Fourier transform. For instance, one analyses oscillations along the temporal axis by applying the Discrete Fourier Transform (DFT) independently to each row of $\X$
\begin{equation}\label{eq:dft}
    \DFT{\X} = \X \bar{\UT},
\end{equation}
where $\UT$ is the normalized DFT matrix defined as
\begin{equation}\label{eq:dftmtx}
    \UT^{\hermitian}(t,k)= \frac{\eu^{-\ju \kk t}}{\sqrt{T}}, \quad \text{with} \quad \kk = \frac{2 \pi (k-1)}{T},
\end{equation}
with $t,k = 1, 2, \ldots, T.$ 
Similarly, the Graph Fourier Transform (GFT)~\cite{hammond2011wavelets,shuman2013emerging,shuman2013vertex} allows us to analyze oscillations along the graph edges. As each column of $\X$ represents a time instant, the GFT of $\X$ for all $t$ reads
\begin{equation}
\GFT{\X} = \widetilde{\X}=\UG^\hermitian\X,
\end{equation}
where $\UG$ is obtained by the eigendecomposition $\LG = \UG\bLambda_G \UG^\hermitian$ of the graph Laplacian. As $\UG$ is orthonormal, the inverse Fourier transform becomes ${\IGFT{\widetilde{\X}} = \X= \UG\widetilde{\X}}$.
This spectral decomposition gives rise to a graph-specific notion of frequency as their squared modulus corresponds the Laplacian eigenvalue $\bLambda_G(\ell,\ell)=\ll$. 

Harmonic time-vertex analysis amounts to analyzing oscillations \textit{jointly} along both the time and the vertex dimensions. Hence, assuming a non-varying graph in time, the \textit{joint time-vertex Fourier transform}, or JFT for short, is obtained by applying the GFT on the graph dimension and the DFT along the time dimension~\cite{loukas2016frequency}
\begin{equation*}
    \widehat{\X}(\ell,k) = \dfrac{1}{\sqrt{T}}\sum_{n=1}^N\sum_{t=1}^{T}\X(n,t)\u_{\ell}^{\hermitian}(n)\eu^{-\ju \kk t}.
\end{equation*}
The above expression can be conveniently rewritten in matrix form as
\begin{equation}\label{eq:jft}
    \Xhat =\JFT{\X} = \UG^* \X \overline{\U}_T.
\end{equation}
Expressed in vector form, the transform becomes 
\begin{equation}\label{eq:vecjft}
\xhat = \JFT{\x} = \UJ^* \x,
\end{equation}
where $\UJ = \UT\otimes\UG$ is the Kronecker product of the basis. 
The relation between Eq.\eqref{eq:jft} and Eq.\eqref{eq:vecjft} is obtained through the property of the Kronecker product
$\lrpar{\M_1\otimes\M_2}\x = \M_2\X\M_1^{\transpose}.$

\begin{table}[t]
\begin{mdframed}
\noindent \textbf{Properties of JFT.}

\vspace{2mm}\noindent Property 1. \textit{JFT is an invertible transform. The inverse JFT in matrix and vector form are, respectively,} $\IJFT{\Xhat} = \UG \X \UT^\transpose $ \textit{and} $\IJFT{\xhat} = \UJ \x$.

\vspace{2mm}\noindent Property 2. \textit{The Parseval relation holds:}
    \begin{equation}\label{eq:parseval}
        \sum_{n,t=1}^{N,T}\abs{\X(n,t)}^2 = \sum_{\ell,k=1}^{N,T}\abs{\Xhat(\ell,k)}^2.
    \end{equation}
    
\vspace{2mm}\noindent Property 3. \textit{The transform is independent on the order GFT and DFT are applied to the time-vertex signal}
    \begin{equation*}
        \JFT{\X} = \GFT{\DFT{\X}} = \DFT{\GFT{\X}}.
    \end{equation*}
    
\vspace{2mm}\noindent Property 4. \textit{The subspace of zero graph and temporal frequency is spanned by the constant time-vertex signal $\boldsymbol{1}\boldsymbol{1}^\hermitian$, with $\boldsymbol{1}$ the all-ones vector.}    
\end{mdframed}
\vspace{-4mm}
\end{table}

\subsection{Time-vertex calculus and variation}
In the following, we briefly present the main time-vertex differential operators. These will help us (a) to perform calculus on a finite, discrete time and space, and (b) to characterize the properties of the signals, such as smoothness, while taking into account the intrinsic structure of the data domain.

\paragraph{Time and vertex domains.} Before introducing the time-vertex operators, we momentarily diverge by presenting the standard definitions in the time and graph domains. The main discrete calculus operator in time is the first order difference operator  
 \begin{equation*}
     \left.\X \GT\right\rvert_t = \x_t-\x_{t-1},
 \end{equation*}
taken here with periodic boundary conditions.
Hence, the symmetric time Laplacian matrix $\LT=\GT^{\hermitian}\GT$ is the discrete second order derivative in time with reversed sign
\begin{equation}\label{eq:tlap}
     \left.\X \LT\right\rvert_t = -\x_{t+1}+2\x_t-\x_{t-1}
\end{equation}
with $\x_{t+1} = \x_1$.
As a circulant matrix, it has eigendecomposition $\LT =\UT \bLambda_T \UT^{*}$, where
\begin{equation}\label{eq:dctfreq}
    \bLambda_T(k,k) = 2\left(1-\cos\left(\kk \right)\right).
\end{equation}

The operator corresponding to the time derivative in the vertex is the edge derivative. Given a graph signal ${\x\in\Rbb^{N}}$, the edge derivative with respect to edge ${e = (n, m)}$ at vertex $n$ is given by
\begin{equation}
\left.\parder{\X}{e}\right\lvert_{n} = \sqrt{\W(n,m)}\lrsquare{\x^n-\x^m}.
\end{equation}
Therefore the graph gradient of $\x$ at vertex $n$ is 
\begin{equation}\label{eq:Ggrad}
\left.\nabla_G \x\right\lvert_{n} = \lrbrace{\left.\parder{\x}{e}\right\lvert_{n}}_{e\in\Es}
\end{equation}
and, as before, $\LG = \divG\GG$, where $\divG$ is the divergence operator of the graph.

\paragraph{Joint domain.} 
We define the joint gradient of a time-vertex signal $\X$ by concatenation of the time and graph gradients:
\begin{equation}\label{eq:jdelx}
    \GJ \x = \vec{\left[\begin{array}{c}
    \X\GT^{\transpose}\\
    \GG \X
    \end{array}\right]}.
\end{equation}
Therefore $\GJ$ can be rewritten as
\begin{equation}\label{eq:joint_gradient}
    \GJ = \left[\begin{array}{c}
    \GT\otimes\I_G\\
    \I_T\otimes\GG
    \end{array}\right].
\end{equation}
The Laplacian is classically defined to equal the divergence of the gradient, and also in our case the joint Laplacian is $\LJ = \GJ^{\hermitian}\GJ$. Expanding the expression while exploiting the mixed-product property of the Kronecker product, we find 
\begin{align*}
    \LJ &= (\GT\otimes\I_G)^\hermitian (\GT\otimes\I_G) + (\I_T\otimes\GG)^\hermitian (\I_T\otimes\GG)\\
    &= (\GT^\hermitian \otimes\I_G)(\GT\otimes\I_G) + (\I_T\otimes\GG^\hermitian) (\I_T\otimes\GG)\\
    &= (\GT^\hermitian\GT)\otimes\I_G + \I_T\otimes(\GG^\hermitian\GG)\\
    &= \LT\otimes \I_G + \I_T \otimes \LG = \LT\times\LG,
\end{align*}
and therefore $\LJ$ is also equivalent to the Cartesian product between the time and  graph Laplacians\footnote{In this work we consider the Cartesian product for its amenable spectral properties, but in general other graph products could be considered, such as the Kronecker product $J = G\otimes G_T$ or the strong product $J = G\boxtimes G_T  = G\times G_T + G\otimes G_T$~\cite{sandryhaila2014big}.}. Above, and the second equality follows from the mixed-product property of the Kronecker product. 
 Thus, the Laplacian operator $\LJ$ applied to the signal $\x$ is
\begin{equation*}
 \LJ\x=\lrpar{\LT\times \LG}\x=\vec{\LG \X + \X\LT}.
\end{equation*}
The result of the Cartesian product is a multilayer graph, referred to as the \emph{joint graph} $J$, where the original graph $G$ is copied at each time step $t = 1,2,\dots , T$. Additionally, each node at layer $t$ is connected to itself at layer $t-1$ and $t+1$ with periodic boundary conditions.
It is useful to remind that the Kronecker product of the two eigenvectors basis $\UT$ and $\UG$ diagonalize the joint Laplacian with eigenvalues equal to the sum of all the pairs $(\kk,\ll)$\cite{merris1994laplacian}:
\begin{align*}
\LJ 
&=  \LT\otimes \I_G + \I_T \otimes \LG\\ 
&= (\UT \bLambda_T \UT^{*})\otimes \I_G + \I_T\otimes (\UG\bLambda_G \UG^{\hermitian})\\
&= (\UT\otimes \UG)(\bLambda_T \times \bLambda_G)( \UT \otimes \UG)^{\hermitian} = \UJ\bLambda_J \UJ^{\hermitian}
\end{align*}
where we have used the mixed-product property of the Kronecker product.

\paragraph{Measures of joint variation.}

The gradient and its various norms are often used as regularizers in regression because they capture the variation of the signal over a domain of interest. The $\ell_2$-norm of the joint gradient measures the total variation of the signal across edges and consecutive steps. Observe that   
\begin{align}\label{eq:tik}
    \| \grad_J \x\|^2_2 = \x^{\transpose}\LJ\x 
    &= \norm{\GG\X}_F^2 + \norm{\X\GT}_F^2 \notag \\
    &\hspace{-15mm}= \trace{\X^{\transpose}\LG\X} + \trace{\X\LT\X^{\transpose}}
\end{align}
meaning that $\| \grad_J \x\|^2_2$ is separable over the the two domains.

Analogously, the $\ell_1$-norm of the joint gradient can be written as the sum of the $\ell_1$-norms
\begin{equation}\label{eq:l1_norm}
\norm{\GJ\x}_1 =  \norm{\vec{\GG\X}}_1 + \norm{\vec{\X\GT}}_1,
\end{equation}
which is often referred as the Total Variation (TV) norm.
In general, it is possible to define a mixed norm $\jnorm{\cdot}{p}{q}$
\begin{equation}\label{eq:lpq_norm}
\jnorm{\x}{p}{q} \triangleq   w_G\norm{\vec{\GG\X}}_p^p + w_T \norm{\vec{\X\GT}}_q^q
\end{equation}
where the $p$-norm and the $q$-norm are computed independently on the two domains and $w_G, w_T$ are non-negative weights. Such norms are often useful when the signal vary differently (e.g., smooth or piece-wise) across the two domains, as we will show in Section~\ref{ssec:inpainting}.

\section{Dynamics over Graphs}\label{sec:DoG}

This section motivates the JFT further by showing that it can be used to characterize two linear PDEs evolving over the graph by kernels defined in the joint frequency domain, and also to provide insight on standard non-linear PDEs used in epidemic modeling.
Our interest in PDEs analysis is related to their capability of encoding information about the structure of the underlying domain, whether continuous or discrete \cite{solomon2015pde,courant1967partial}
Moreover, PDEs are not only simple but powerful models of many phenomena observed in reality, but also a motivation for the Fourier transform~\cite{fourier1807memoire,narasimhan1999fourier}.

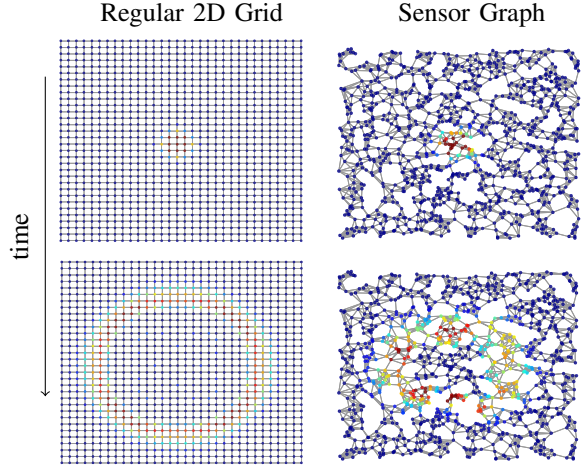
\begin{figure}
\centering
\begin{minipage}{0.05\linewidth}
\begin{flushleft}
\begin{tikzpicture}
\node [guide] at (0,1) (a) {};
\node [guide] at (0,-3.5) (b) {};
\draw[->] (a) --node[sloped, above=-0.1cm,rotate=180] {time} (b);
\end{tikzpicture}
\end{flushleft}
\end{minipage}
\begin{minipage}{0.93\linewidth}
\centering
\hspace{-0.1cm} Regular 2D Grid	\qquad	 \qquad		Sensor Graph\\
\includegraphics[width=0.47\linewidth]{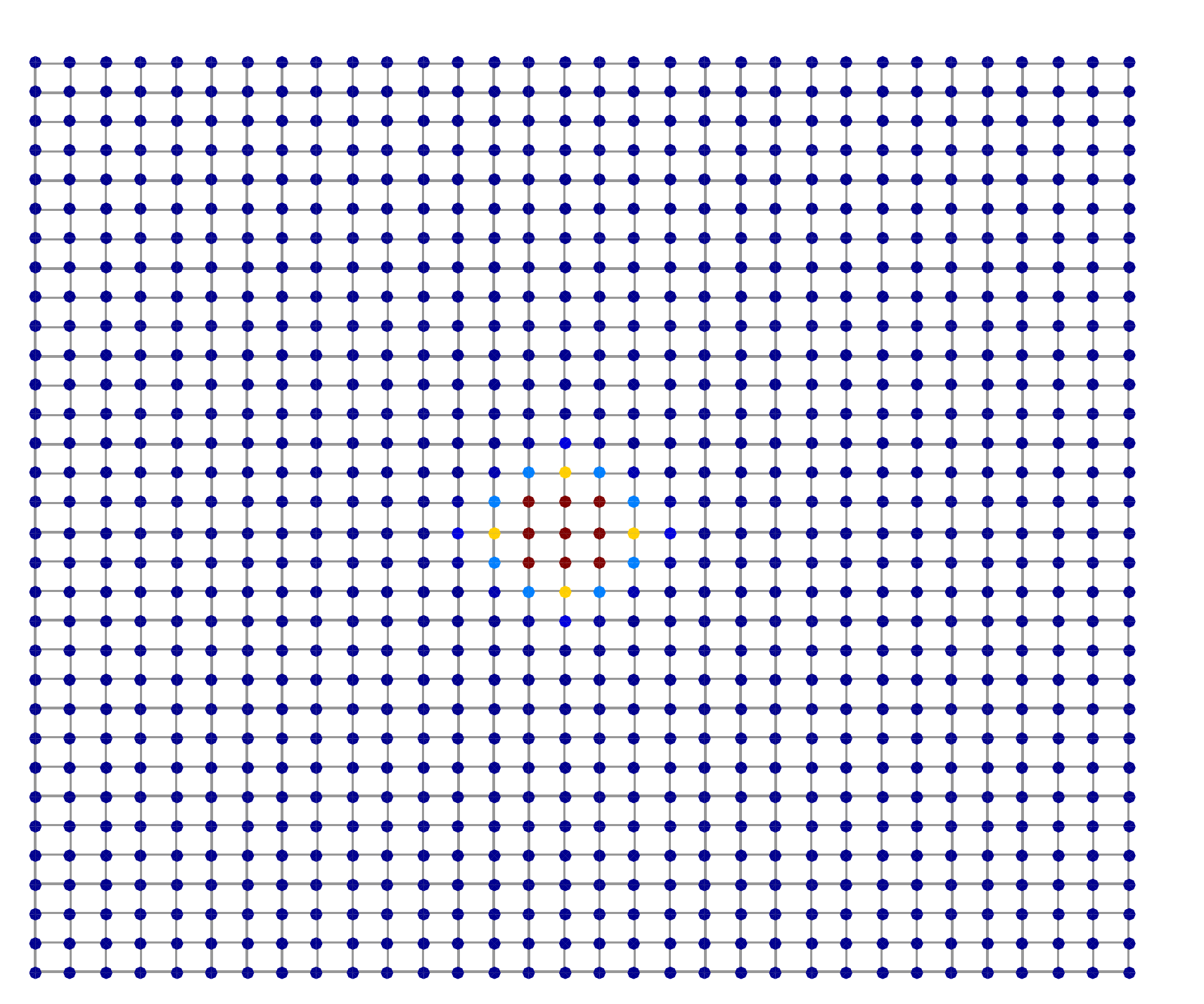} \includegraphics[width=0.47\linewidth]{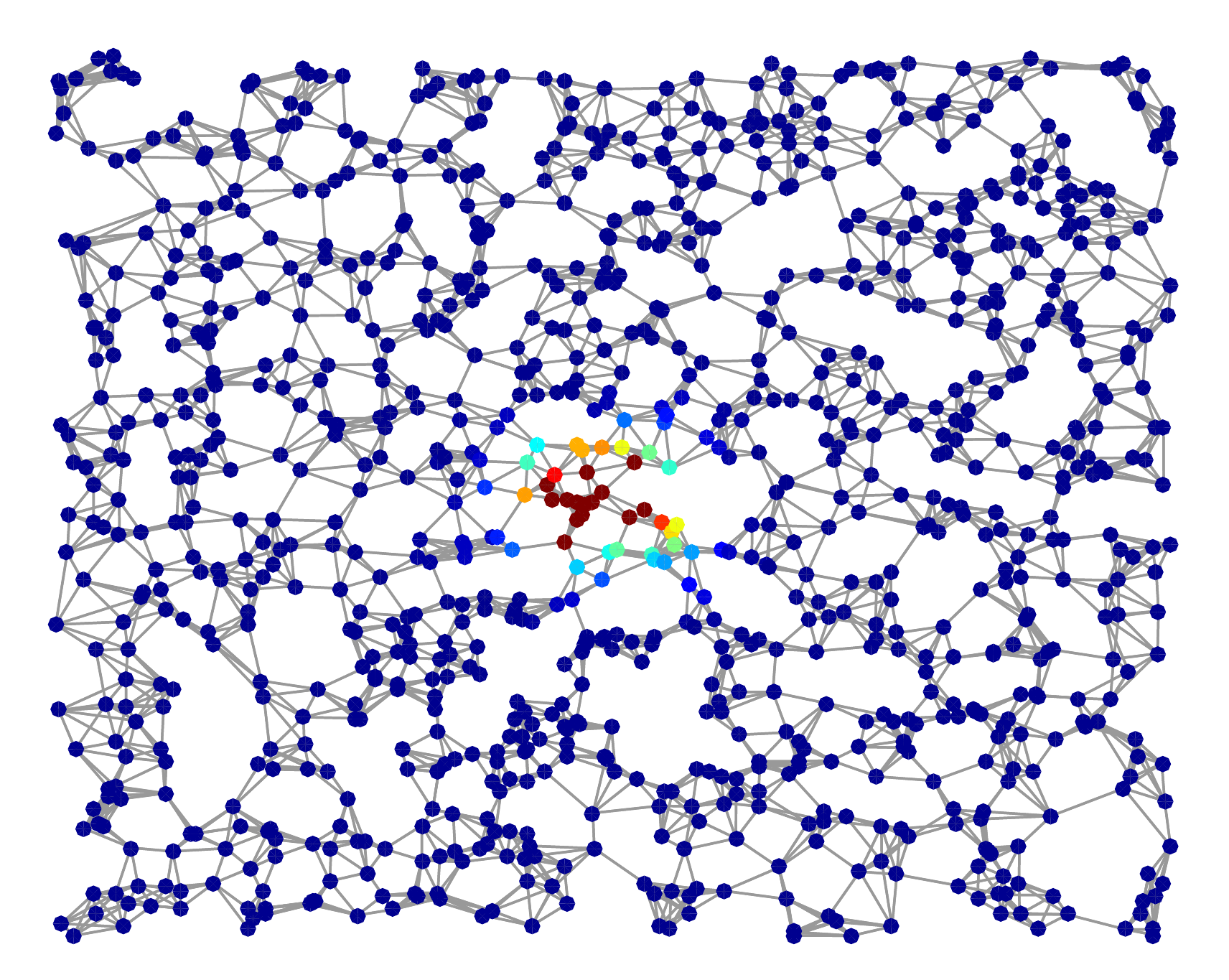}\\
\includegraphics[width=0.47\linewidth]{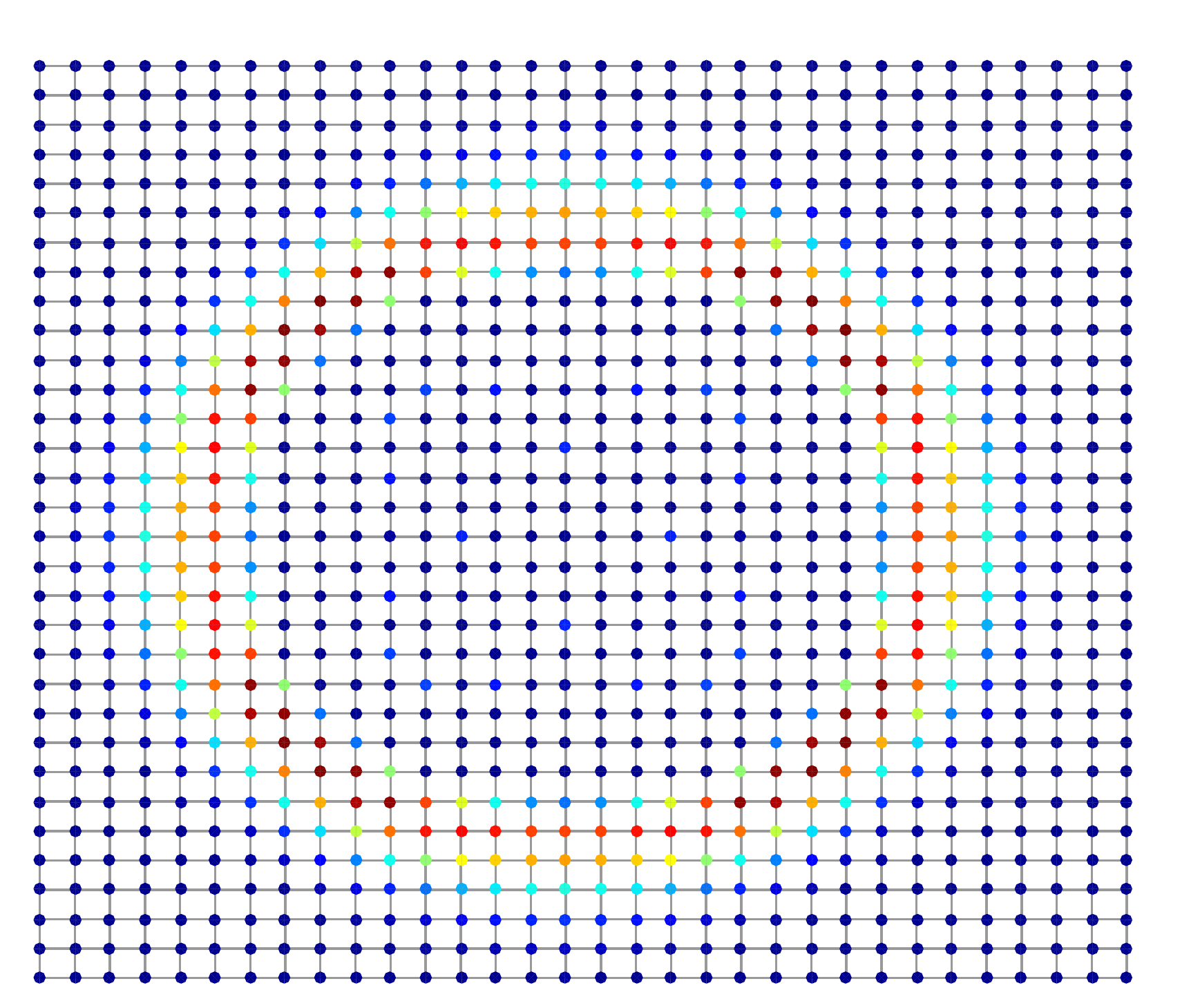} \includegraphics[width=0.47\linewidth]{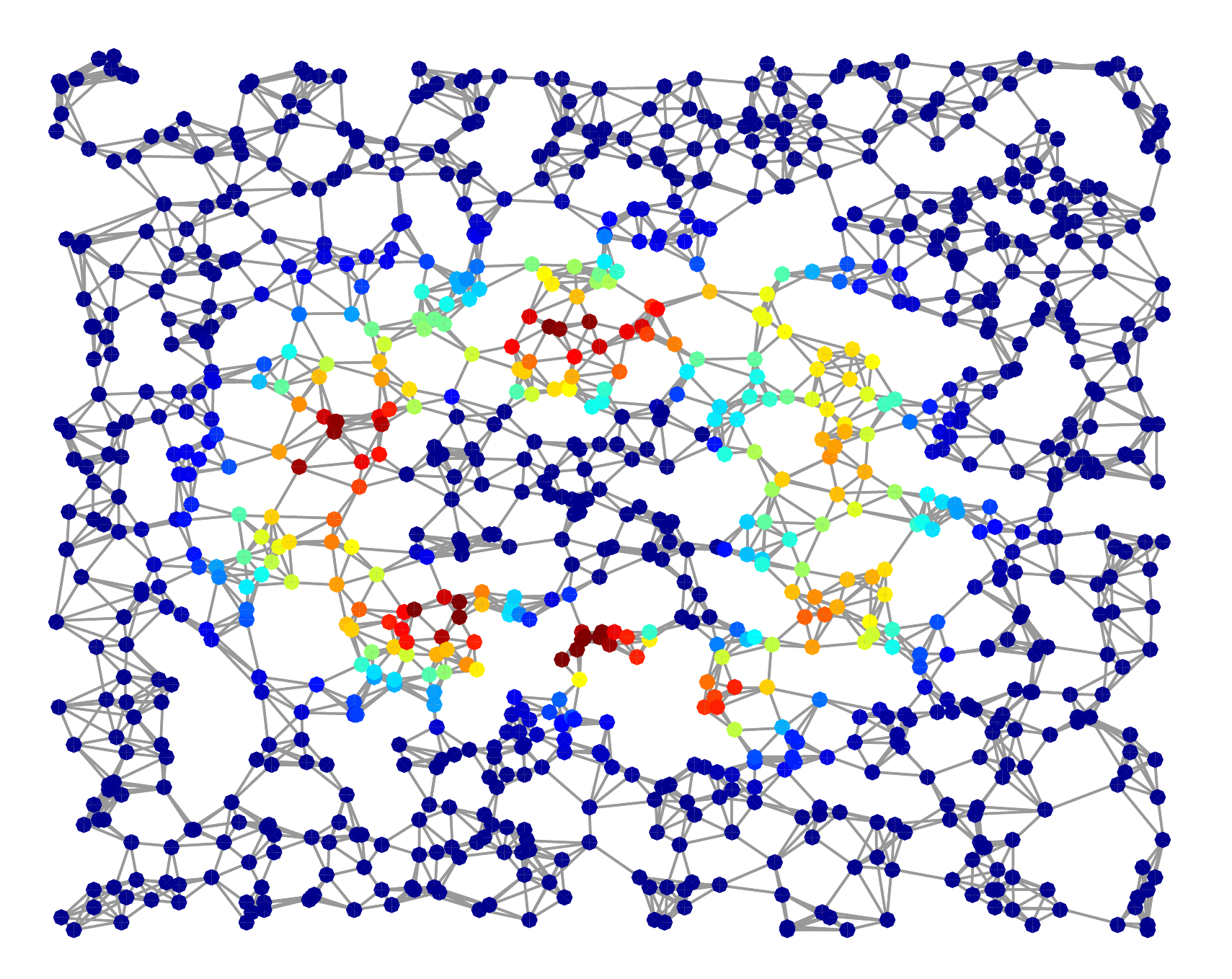}
\end{minipage}
\caption{Solution to the wave equation on a regular 2D grid and on a sensor graph at different point in time. The propagating behavior is evident even in the case of irregular domain.}\label{fig:wave_example}
\end{figure}

\subsection{Linear dynamics on graphs}\label{ssec:linearPDE}
We are interested in linear PDEs whose solution at each time step can be expressed as a linear operator applied to the initial condition. In particular, we consider the heat diffusion and the wave equations defined in the discrete setting. We denote $\x_1$ the initial condition of the PDEs, or equivalently in the joint spectral domain ${\Z({\ell,k}) = \widetilde{\x_1}(\ell) \, \U_T^*(k,1)}$.

\paragraph{Heat equation.} The discrete heat diffusion equation ${\x_t-\x_{t-1} = -s  \LG \x_t}$ is, arguably, one of the simplest dynamics described by differential equations. The parameter $s$ represents thermal diffusivity and is interpreted as a scale parameter for multiscale dynamic graph wavelet analysis~\cite{coifman2006diffusion} and graph scale-space theory~\cite{loukas2015graph}. 
It is well understood that 
\begin{equation}\label{eq:heat}
    \x_t = (\I - s\LG )^{t-1} \x_1.
\end{equation}
Evaluating both the GFT and DFT, one also finds that the solution also has distinct structure in the joint spectral domain
\begin{equation}
    \Xhat(\ell,k) = \frac{1}{\sqrt{T}}\frac{a(\ll,\kk)^T - 1}{a(\ll,\kk)-1} \Z(\ell,k)
\end{equation}
where $a(\ll,\kk) = (1 - s\ll) \, \eu^{-\ju \kk}$. The JFT of a heat diffusion process therefore exhibits a smooth non-separable low-pass form.

\paragraph{Wave equation} 
More interesting dynamics can be modeled by the discrete second order differential equation 
\begin{equation}\label{eq:wavepde2}
\X \LT=s \LG \X,
\end{equation}
representing a discrete wave propagating on a graph with speed $s>0$. Figure~\ref{fig:wave_example} shows the evolution of a signal obtained by solving~\eqref{eq:wavepde2} using a numerical iterative scheme on a regular 2D lattice and on a random sensor graph. It is clear that, assuming a sufficiently regular graph, the solution to Eq.~\eqref{eq:wavepde2} evolves on the graph as a wave propagates in the Euclidean domain. In the appendix we prove that, for vanishing initial velocity, the solution in the joint spectral domain can be written as
\begin{equation*}
    \Xhat(\ell,k) = \widehat{K}_{s}(\ll,\kk)\Z(\ell,k)
\end{equation*}
where
\begin{equation}\label{eq:wave_ker}
\widehat{K}_{s}(\ll,\kk) =	\sum_{t}\cos(t\tl)\eu^{-\ju \kk t}
\end{equation}
and $\tl = \arccos(1-\frac{s\ll}{2})$.
Since the $\arccos(x)$ is defined only for ${x\in\lrsquare{-1,1}}$, to guarantee stability the parameter $s$ must satisfy ${s<4/\lambda_{max}}$.

\begin{figure}[t!]
\centering
\includegraphics[width=\columnwidth]{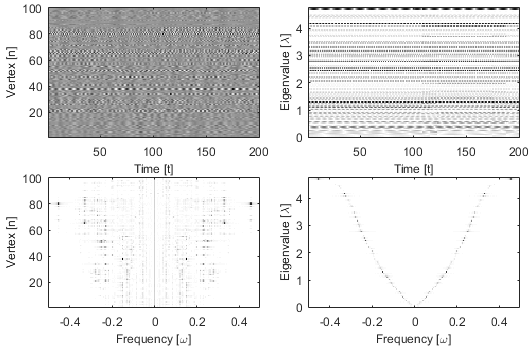}
\caption{Frequency analysis of multiple waves propagating on a random sensor graph. The signals on each node of the graph evolve according to a PDE, but their time-vertex representation (top left) does not highlight any relation between the two domains. Similarly, GFT (top right) and DFT (bottom left) are not able to show the underlying structure. It is evident that JFT (bottom right) succeeds in representing the signal in the most meaningful way, revealing its regular pattern.
}\label{fig:wavespectra}
\end{figure}

The distinctive pattern of the JFT of a wave shown in Fig.~\ref{fig:wavespectra} (bottom right) is sparser and more structured that the original (top left), GFT (top right), and DFT (bottom left) representations. 

\subsection{Complex dynamics over networks: the illustrative example of epidemic models}
\label{ssec:epidem}

Time-vertex harmonic analysis often provides useful insights on the dynamics of a signal even when the latter is not characterized by a linear PDE. To illustrate this, we will show how the JFT can be used to characterize the evolution of a non-linear, discrete, and non-deterministic model for the spread of an infectious disease.

In particular, we focus on the dynamics corresponding to different compartmental models commonly used in epidemiology. We simulated the epidemics spreading over N = 695 cities of Europe according to two different models: the Susceptible-Exposed-Infected-Recovered (SEIR) model and the SEIRS model, where the immunity of recovered individuals is only temporary. The models are parametrized by the contagion probability of infection, the infectious, latent and immunity periods and the population per city. Each node of the graph represents a city with a fixed population of individuals. Connections within the cities are modeled using randomly generated Erd\H{o}s-R\'enyi graphs. Inter-cities connection are modeled using two graphs, a terrestrial coordinate-based graph and the graph of airline connections between the major city in Europe.

\begin{figure}[t!]
\includegraphics[width=\linewidth]{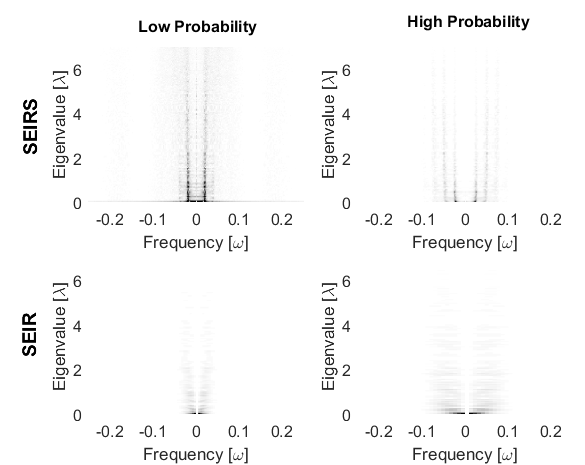}
\caption{JFT for different realization of epidemic spreads, modeled using different models and contagion probability. The transform allows us to distinguish between the different parameters of the model. }\label{fig:epid_spectra}
\end{figure}

We simulate different realizations of the epidemic breakout, varying the parameters and observing how they influence the joint spectrum of the signals describing the evolution of the number of infected individuals at each node. Figure~\ref{fig:epid_spectra} shows the JFT of those signals. In the first case, the spectrum is characterized by regularly spaced lines along the angular frequency axis, caused by the inherent periodicity of the SEIRS model, where every individual can be infected again after the temporary immunity period ceases. The spacing between the lines is due to both the immunity and latent periods. In the second case the epidemic has a more diffusive behaviour, without evident periodicity. For each one of the models, we simulate scenarios with high and low probabilities of contagion. It can be observed that the high probability case is characterized by a larger bandwidth occupancy with respect to the low probability one, due to a more impulsive behaviour of the epidemic breakout. Therefore, JFT seems to reveal the underlying structure and allows us to differentiate between the different models.

\section{Fast Filtering of Time-Vertex Signals} \label{sec:Jop}

After recalling the definition of joint filters, we next present a novel algorithm to perform fast filtering on large graphs. Experiments illustrate that our algorithm achieves significantly better approximation of filtering objectives than state-of-the-art, while also not being constrained to a specific class of separable responses.     

\subsection{Joint time-vertex filters}

A {joint filter} $h(\LG,\LT)$ is a function defined in the joint two-dimensional spectral domain ${h: \Rbb_+ \times \Rbb \mapsto \Cbb}$ and is evaluated at the graph eigenvalues $\lambda$ and the angular frequencies $\omega$. Similarly to both the classical and the graph case, the generalized filtering operator is applied by means of the convolution theorem, \ie point-wise multiplication in the spectral domain. The output of a joint filter is
\begin{align} \label{eq:joint_filtering}
    h(\LG,\LT) \, \x &= \UJ\, h(\bLambda_G,\bOmega) \, \UJ^\hermitian \x,
\end{align}
where $h(\bLambda_G,\bOmega)$ is a diagonal $NT\times NT$ matrix defined as  
\begin{align}
     h(\bLambda_G,\bOmega) = \diag{\begin{bmatrix}
 h(\lambda_1, \omega_1) & \cdots & h(\lambda_1, \omega_T) \\
 \vdots &  \ddots & \vdots \\
 h(\lambda_N, \omega_1)  & \cdots & h(\lambda_N, \omega_T)
 \end{bmatrix}} \notag 
\end{align}
and the $\diag{\A}$ operator creates a matrix with diagonal elements the vectorized form of $\A$.

\begin{figure}[t!]
\centering
    \includegraphics[width=0.65\columnwidth]{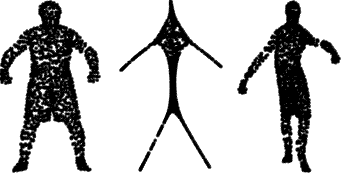}
    \caption{The effect of joint filters is easily visualized for the case of a dynamic mesh of a dancer. By filtering the original mesh (left) using a joint low-pass separable filter one approximates the time-varying skeleton of the dancer (center). Using a non-separable wave filter, the fluidity of the dancer's motion is emphasized (right).}
    \label{fig:dancer_filters}
\end{figure} 

\paragraph{Illustration: dynamic mesh filtering.} Figure~\ref{fig:dancer_filters} shows an example of joint filtering of a mesh representing a dancer\footnote{\url{http://research.microsoft.com/en-us/um/redmond/events/geometrycompression/data/default.html}}. We design (a) a joint separable lowpass filter that attenuates high frequency components in both graph and time domains, and (b) a wave filter whose frequency response is described in Eq.~\eqref{eq:wave_ker}. In the first case, we obtain the approximate skeleton of the mesh with rigid movements, whereas the wave filter produce a fluid (wavy) dancer, enhancing the frequency components in a non-linear fashion. We remark that this effect can only be obtained using non-separable filters.

\paragraph{Separable vs. non-separable filters.} A notable family of joint filters are those that have \textit{separable} response \begin{equation}\label{eq:fseparable}
h(\lambda,\omega) = h_1(\lambda) \, h_2(\omega).
\end{equation}
These filters have a straightforward interpretation: the frequency response of each filter affects only the domain where it is defined
\begin{equation}
    h(\LG,\LT) \x  = \vec{h_1(\LG)\X h_2(\LT)}.
\end{equation}
Moreover, since they can be designed independently at the two domains, joint filters can be obtained by combining graph and temporal filters~\cite{isufi2016separable}.
However, due to their simple form, separable filters cannot model the dynamics of PDEs (e.g., waves or heat diffusion), where there is  an interplay between the temporal and graph frequency domains. For this reason, in the following we aim to find an efficient joints filter implementation for separable as well as non-separable filtering objectives.

\subsection{Fast joint filtering}
\label{subsec:FFC}
Due to the high complexity of eigendecomposition, graph filters are almost always implemented using fast 2D polynomial~\cite{shuman2011chebyshev} and rational~\cite{isufi2017autoregressive} approximations. In the context of time-vertex analysis, the importance of fast joint filtering is emphasized by the increase of the problem's dimensions.  
Recognizing this need, researchers have recently proposed distributed joint filter 2D Chebychev polynomial~\cite{loukas2016frequency} and separable rational~\cite{isufi2016separable} implementations, appropriate for arbitrary and separable joint response functions, respectively. 
In the following, we improve upon state of the art by enhancing the filtering approximation at a similar (up to logarithmic factors) complexity. 

\paragraph{The Fast Fourier Chebyshev (FFC) algorithm.} The basic idea of our algorithm is to exploit the small complexity of FFT and perform graph filtering in the time-frequency domain. Concretely, to filter $\X$ with response $h(\lambda,\omega)$, we do the following:
\begin{enumerate}
    \item[1.] Compute the FFT of every row of $\X$, at a total complexity of $\mathcal{O}(NT\log{T})$.
    \item[2.] For each $\kk$, approximate $h(\lambda,\kk)$ with a Chebyshev polynomial of order $M_G$ and use the fast graph Chebyshev recursion~\cite{hammond2011wavelets} to filter the corresponding angular frequency component of $\X$. The complexity of this step is $\mathcal{O}(M_G T\abs{\Es})$.
    \item[3.] Use the inverse FFT to obtain the filtered time-vertex signal, with complexity $\mathcal{O}(NT\log{T})$.
\end{enumerate}
Our scheme can approximate both separable or non-separable joint filters
using $\mathcal{O}( T\abs{\Es}M_G + NT\log{T})$ operations, which up to a logarithmic factor is a linear complexity to the number of edges $\abs{\Es}$, nodes $N$, timesteps $T$, and filter order $M_G$. Moreover, it can be performed distributedly since both the FFT and the graph Chebychev recursion necessitate only local or few hop information. 

\begin{figure}[t]
\includegraphics[width=\columnwidth]{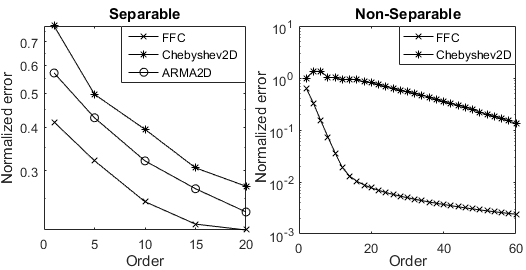}
\caption{
Fast joint filtering comparison using different algorithms to approximate the ideal joint lowpass filter (left) and a non-separable wave filter (right) approximated in Eq~\eqref{eq:lpapprox} and Eq~\eqref{eq:waveapprox}, respectively. The proposed method (FFC) outperforms the others, in particular for non-separable filters.}\label{fig:filter_approx}
\end{figure}

\paragraph{Numerical comparison.} To evaluate the approximation properties of the above scheme, we show in Figure~\ref{fig:filter_approx} numerical experiments for an ideal separable lowpass filter and a non-separable wave filter on a time-vertex graph with size $N = 5000$, $T = 3000$. In detail, the approximated filtering functions (low pass and wave) are, respectively,
\begin{align}
    h_{\mathrm{LP}}(\ll,\kk) &= \dfrac{\eu^{-(\ll-\lambda_{\mathrm{cf}})}}{1+\eu^{-(\ll-\lambda_{\mathrm{cf}})}} \dfrac{\eu^{-(\abs{\kk}-\omega_{\mathrm{cf}})}}{1+\eu^{-(\abs{\kk}-\omega_{\mathrm{cf}})}}\label{eq:lpapprox}\\[1em]
    h_{\mathrm{wave}}(\ll,\kk) &= \eu^{-\abs{\pi\abs{\kk}-\arccos(1-\ll/(2\lmax)}^2}.\label{eq:waveapprox}   
\end{align}
For each case, we compare our algorithm with the state-of-the-art, i.e., Chebyshev2D approximation~\cite{loukas2016frequency} of complexity and the ARMA2D approach~\cite{isufi2016separable}, while choosing $M_G$ and $M_T$ as graph and temporal polynomial orders, respectively (here $M_G=M_T$). 

As shown in Figure~\ref{fig:filter_approx}, FFC results in a significant improvement in accuracy for the same order and the difference is particularly prominent in the non-separable case (ARMA2D cannot be used here). We remark however that, to interpret these results correctly, one has to consider the complexity of each method:

\setlength{\arrayrulewidth}{0.7pt}
\setlength{\tabcolsep}{4pt}
\renewcommand{\arraystretch}{1.5}

\begin{center}
\small
\begin{tabular}{ r l c}
 method & complexity & applicability \\ \hline  
 \textbf{FFC} & $\mathcal{O}(T\abs{\Es} M_G + NT\log{T})$ & all \\ 
 Cheby2D~\cite{loukas2016frequency} & $\mathcal{O}(T\abs{\Es} M_T + NT M_T M_G)$ & all \\  
 ARMA2D~\cite{isufi2016separable} & $\mathcal{O}(T\abs{\Es}M_G + T\abs{\Es}M_T)$ & separable
\end{tabular}
\end{center}
Therefore, for the same order, the three different methods feature slightly different complexities, implying that a direct comparison of accuracy is not entirely fair. Nevertheless, the unfairness in not in our favor as, in our experiments for all orders larger than 2, the asymptotic complexity of FFC is the smallest (since here $M_G=M_T$, $\log{T} < M_T M_G$, and $\log{T} < \abs{\Es} M_T/N$).     
We also note that, in practice one often needs $M_T \gg \log{T}$ to achieve a good approximation, in which case FFC is the fastest.

\section{Time-vertex Dictionaries and Frames} \label{sec:Jframes}

So far, we have looked at time-vertex signals through the lenses of the canonical and the joint Fourier bases. However, in some cases it is beneficial to also consider alternative representations. For example, in the classic case, the wavelet and the short time Fourier transforms respectively enable time-scale and time-frequency analysis of the signal. The purpose of this section is to show how one can define analogous representations for time-vertex signals. These can be used for instance to generate features given as an input to a classifier (see Section~\ref{ssec:clustering}) or to regularize an optimization problem such as \eqref{eq:opt_l1_dict} in  Section~\ref{ssec:source}.

Classically, the atoms of the representations are built by applying a transform (e.g. scaling or modulation) to a mother function and shifting the resulting functions. We follow a similar approach, with the difference that the mother function is replaced by a kernel defined in the time-vertex frequency domain and the shifting has to be replaced by an operator suitable to graphs. The spectral time-vertex wavelet and the short time vertex Fourier transforms follow as consequences of our framework.

\subsection{Joint time-vertex localization}
The ability to localize a kernel over a particular time and vertex is a key ingredient of our dictionary construction. In the following, we derive such a {joint localization operator} as a generalization of the {graph localization operator}~\cite{perraudin2016global,perraudin2016stationary,shuman2013vertex}, which localizes a kernel $h(\LG)$ onto vertex $v_{m}$
\begin{equation}
\label{def:localization_operator}
    \TG{h}{m} \delequal h(\LG) \, \b{\delta}_{m}= \sum_{\ell=1}^{N} h(\ll)\, \bar{\u}_{\ell}(m)\, \u_{\ell}.
\end{equation}
Above, $\b{\delta}_{m}$ is a Kronecker delta centered at vertex $v_{m}$. 
Similarly, in the joint domain we define the \emph{joint time-vertex localization operator} as the filtering with a two-dimensional Kronecker delta
\begin{align} 
    \TJ{h}{m,\tau}  
    &\delequal   \, h(\LG,\LT) \, \lrpar{\b{\delta}_{m} \otimes \b{\delta}_{\tau}}. \label{eq:joint_localization_mat}
\end{align}
It turns out that the joint time-vertex localization operator has the advantages of both the graph localization and the traditional translation operator. 
Indeed, we observe the following relations 
\begin{align}
 \TJ{h}{m,\tau}(n,t) 
    &= \frac{1}{T} \hspace{0mm} \sum_{\substack{\ell=1\\k = 1}}^{N, T} \hspace{0mm} h(\ll,\kk)\bar{\u}_{\ell}(m)  \eu^{-\ju\kk\tau} \u_{\ell}(n)\eu^{\ju\kk t}\nonumber \\
    &\hspace{-17mm}= \frac{1}{T} \hspace{0mm} \sum_{k = 1}^{T}\left(\sum_{\ell=1}^N \hspace{0mm} h(\ll,\kk)\bar{\u}_{\ell}(m)\u_{\ell}(n) \right) \eu^{\ju\kk(t-\tau)}  \label{eq:joint-translation-property} 
\end{align}
From \eqref{eq:joint-translation-property}, it follows that joint localization consists of three steps:
(a) localizing independently all kernels $h(\cdot, \omega_k)$, (b) computing the inverse DFT along the other dimension, and (c) translating the result. Joint localization is thus equivalent to independent application of a graph localization and a translation.
Note that the steps (a) and (b) can be considered as localizing the signal along the time dimension.

When the filter is separable, i.e. $h(\lambda,\omega) = h_G(\lambda)h_T(\omega)$, the joint localization is simply
\begin{equation}
    h(\LG,\LT) \, (\b{\delta}_{\tau} \otimes \b{\delta}_{m})  = \vec{ h_G(\LG)  (\b{\delta}_{\tau} \otimes \b{\delta}_{m}^\transpose) h_T(\LT)},
\end{equation}
showing that the filter can be localized independently in time and in the vertex domain.

\subsection{Joint time-vertex dictionaries} \label{sec:JTVD}

We proceed to present our dictionary construction for time-vertex signals. We start with a mother time-vertex kernel $h(\lambda, \omega)$ and a transformation $s_{z_\lambda,z_\omega}(\cdot, \cdot)$ parametrized by some values $[z_\lambda,z_\omega]$ belonging to the finite 2D set $\mathcal{Z}_\lambda \times \mathcal{Z}_\omega \subset \Rbb^2$ and controlling the kernel's shape along the vertex and time domains. The transformed kernel is then obtained by composition
\begin{align}
    h_{z_\lambda,z_\omega}(\lambda,\omega) = h(s_{z_\lambda, z_\omega}(\lambda,\omega)).
\end{align}
We build our dictionary by transforming $h(\lambda, \omega)$ with all $ {z_\lambda,z_\omega} \in \mathcal{Z}_\lambda \times \mathcal{Z}_\omega$, (possibly) normalizing, and jointly localizing the resulting kernels $h_{z_\lambda,z_\omega}(\lambda,\omega)$ at each node $m$ and time $\tau$. Concretely, the dictionary is 
\begin{align} 
\mathcal{D}_{h} = \{ \TJ{h_{z_\lambda,z_\omega}}{m,\tau}\} \quad &\text{for} \quad m \in \mathcal{V},\ \tau =1,2, \ldots, T, \notag \\&\hspace{0mm} \text{and} \ {[z_\lambda,z_\omega]} \in \mathcal{Z}_\lambda \times \mathcal{Z}_\omega \}.
\end{align}
When $\mathcal{D}_{h}$ is overly redundant, one may choose to consider only a subset of values for $m$ and $\tau$. 

\vspace{2mm}\noindent We next consider two interesting examples of the proposed dictionary construction that are generalizations of the short-time Fourier and wavelet transforms~\cite{christensen2016introduction,grochenig2013foundations}:

\paragraph{Short Time-Vertex Fourier Transform (STVFT).} Set $s_{z_\lambda,z_\omega}(\cdot, \cdot)$ to a shift in the spectral domain 
\begin{align}
    s_{z_\lambda,z_\omega}(\lambda,\omega) = [\lambda-z_\lambda, \omega-z_\omega].
\end{align}
This transform can be considered as a modulation. Nevertheless, we note that it does not correspond to a multiplication by an eigenvector as in~\cite{shuman2012windowed,shuman2013vertex}. Our construction is more related to~\cite[Section 3]{shuman2015spectrum}.
Then, given a separable mother kernel $h(\lambda,\omega) = h_G(\lambda)\, h_T(\omega)$ and a finite 2D set $\mathcal{Z}_\lambda \times \mathcal{Z}_\omega \subset \Rbb^2$, the STVFT of signal $\X$ is defined as

\begin{small}
\begin{align*}
\STVFT{\X}(m,\tau,z_\lambda,z_\omega)\delequal  \langle \X,\TJ{h(\lambda-z_\lambda,\omega-z_\omega)}{m,\tau}\rangle \\
  = \frac{1}{\sqrt{T}}\sum_{\ell,k}h(\ll-z_\lambda,\kk-z_\omega)\widehat{\X}(\ell,k)\u_{\ell}(m)\eu^{\ju \omega_k \tau}.
\end{align*}
\end{small}

Provided that $h(\lambda, \omega)$ is localized around $[0,0]$, the amplitude of the coefficient $(n,t,z_\lambda,z_\omega)$ indicates the presence of the spectral mode $[z_\lambda,z_\omega]$ at vertex $m$ and time $\tau$. 
Moreover, since the mother kernel is separable, the design in the two domains can be performed independently:

\textit{For the graph domain}, we suggest to select the values of $z_\lambda$ to be equally spaced in $[0,\lmax]$~\cite[Section 3]{shuman2015spectrum}. The spacing should be selected such that $\sum_{z_{\lambda} \in \mathcal{Z}_\lambda}  h_G^2(\ll-z_\lambda) \approx c$ for every $\ll$, ensuring good conditioning of the associated frame (see \cite[Theorem 5.6]{hammond2011wavelets}). Because of the graph irregularity,  in most of the cases, we need to keep all possible values for $m$, i.e., $m=1,2,\dots, N$.

\textit{For the time domain}, we recover a traditional STFT, with the difference that $h_T(\omega)$ is defined in the spectral domain. Nevertheless, for convenience, the window can still be designed in the time domain.
As a rule of thumb $|\mathcal{Z}_\omega| = l_{h_T}$, where $l_{h_T}$ is the support of $h_T$ in the time domain\footnote{In practice, the kernel is chosen to have a compact support in the time-domain.}, and the values of $\tau$ should be sampled regularly with a spacing $\frac{l_{h_T}}{R}$, where $R$ is the desired redundancy. For a more complete treatment we refer the reader to~\cite{christensen2016introduction}.

\paragraph{Spectral Time-Vertex Wavelet Tranform (STVWT).} Following the idea developed in~\cite{hammond2011wavelets}, we
set $s_{z_\lambda, z_\omega}(\cdot, \cdot)$ to a generalized graph dilation (or scaling), i.e., a multiplication in the spectral domain
\begin{align}
    s_{z_\lambda, z_\omega}(\lambda,\omega) = [z_\lambda\,\lambda,\, z_\omega \, \omega].
\end{align}
Then, given a kernel $h(\lambda,\omega)$ the STVWT of $\X$ reads
\begin{align*}
\STVWT{\X}(m,\tau,z_\lambda, z_\omega)\delequal  \langle \X,\TJ{h(z_\lambda\,\lambda,z_\omega \, \omega)}{m,\tau}\rangle \\
  = \frac{1}{\sqrt{T}}\sum_{\ell,k}h(z_\lambda\,\lambda,z_\omega \, \omega)\widehat{\X}(\ell,k)\u_{\ell}(m)\eu^{\ju \omega_k\tau},
\end{align*}
where $z_\lambda, z_\omega$ are the scale parameters for the vertex and the time dimensions. A usual requirement for $h(\lambda,\omega)$ is that it has a zero DC component, i.e., $h(0,0) = 0$. Contrarily to the STVFT, the mother kernel here may not be separable, as illustrated in~\ref{ssec:source}.
The choice of the discretization lattice $[m,\tau,z_\lambda, z_\omega]$ is thus more involved and case dependent: we suggest that $m$ and $\tau$ take all possible $N$ and $T$ values respectively, while $z_\lambda$ and $z_\omega$ are carefully selected depending on the application. This choice is justified by the computational complexity detailed in the following.

\subsection{Joint time-vertex frames}
To make the proposed dictionaries and associated signal representations usable in practice, we next provide answers to three key questions: (a) How can we compute the representations efficiently (i.e., performing analysis and synthesis)? (b) How can we guarantee that the associated transforms are well conditioned such that they can be successfully inverted? (c) How to efficiently invert them, recovering the original signal? 
The second point is particularly important since a well conditioned transform allows for more robust representations, for instance when the dictionary is used to solve a synthesis or analysis regression problem with a sparse regularizer.

\paragraph{Efficient analysis and synthesis.} The dictionary atoms can be seen as a filter-bank $\{h_{z}(\lambda, \omega)\}_{z \in \mathcal{Z}_\lambda\times \mathcal{Z}_\omega}$, in which case the operators going from the signal to the representation domain and back are the \textit{analysis} operator 
\begin{align*}
\Dy{h}\{\X\} (m,\tau,z) = \langle \X,\TJ{h_{z}}{m,\tau}\rangle =  \C_{z}(m,\tau),
\end{align*}
and the \textit{synthesis} operator 
\begin{align*}
\Dy{h}^{\hermitian}\{ \b{C}\} (n,t)
    &= \hspace{0mm}  \sum_{z}\langle \C_{z}, \TJ{h_{z}}{n,t}\rangle  = \Y(n,t).
\end{align*}
Notice that in general $\X \neq \Y$, and equality holds only when the filter-bank is a unitary tight frame. 

Instead of computing the dictionary explicitly (an operation that is costly both in memory and in computations), one may acquire the analysis coefficients for all $m,\tau$ by joint filtering $\X$ with kernel $h_{z_\lambda,z_\omega}$ taking advantage of the relation $\C_{z} = \mat{h_z(\LG,\LT)\x}$. Similarly synthesis can be performed by summing filtering operations. Using our FFC filtering algorithm presented in Section~\ref{subsec:FFC}, the total analysis complexity is thus $\mathcal{O}( |\mathcal{Z}_\lambda \times \mathcal{Z}_\omega| (T
|\mathcal{E}| M_G + NT\log{T}) )$, where typically $M_G \approx 50$.

The drawback of this technique is that it does not allow us to take advantage of sub-sampling the lattice $[n,t]$, especially in the non-separable case. Indeed, a filtering operation will always provide all coefficients regardless of the desired lattice. While addressing this computational problem is beyond the scope of this contribution, we note that we can still efficiently perform a sub-sampled STVFT  by first filtering only in the graph domain and second computing a traditional STFT.

\paragraph{Conditioning and frame bounds.}
In several applications in signal processing one is interested not only in processing data in another convenient representation, but also to recover the original signal from its alternative representation. Redundant invertible dictionaries are referred to as \emph{frames}~\cite{christensen2016introduction,kovacevic2007life1}.
The following theorem generalizes the classic~\cite{christensen2016introduction} results regarding the frame bounds, providing a condition for a joint time-vertex dictionary to be a frame, as in the case of graphs~\cite[Lemma 1]{shuman2015spectrum},~\cite[Theorem 5.6]{hammond2011wavelets}.
\begin{theorem}
\label{theo:frame}
Let $\lrbrace{h_z(\lambda, \omega)}_{z \in \mathcal{Z}_\lambda \times \mathcal{Z}_\omega}$ be the kernels of a time-vertex dictionary $\mathcal{D}_{h}$, and set
\begin{align*}
A=\min_{l,k}\sum_{z}\abs{h_{z}(\ll,\omega_{k})}^{2}, \
B=\max_{l,k}\sum_{z}\abs{h_{z}(\ll,\omega_{k})}^{2}.
\end{align*} 
If $0 < A \leq B < \infty$, then $\Dy{h}$ is a frame in the sense:
\begin{equation}\label{eq:framebound}
A\norm{\X}_F^2\leq \norm*{\Dy{h}\{\X\}}_F^2 \leq B\norm{\X}_F^2
\end{equation}
for any time-vertex signal $\X\in\Rbb^{N\times T}$.%
\end{theorem}
The proof of theorem can be found in the appendix (see section \ref{ssec:proof}).

The theorem asserts that, if $A>0$, no information is lost when the analysis operator is applied to a time-vertex signals, thus the transform is invertible. 
Furthermore, the ratio of the frame bounds $A/B$ is related to the condition number of the frame operator $S_h\{\X\}=\Dy{h}^{\hermitian}\{\Dy{h}\{\X\}\}$, hence it is decisive for efficient reconstruction when we want to recover the signal from its representation solving an optimization problem~\cite[Section 7]{hammond2011wavelets}. 

\paragraph{Efficient inversion.}
To recover the signal $\X$ from the coefficients $\C$, a solution is to use the pseudo-inverse, i.e. $\X = \Dy{h}^\dagger \{\C\}$ or to solve the following convex problem $\argmin_\X \norm{\Dy{h}\{\X\}-\C}_2^2$. Problematically, these are computationally intractable for large value of $N$ and $T$. We will instead design a dual set of kernels that  allows us to invert the transform by a single synthesis operation. To this end, we search for a set of filters $\tilde{h}$ such that $\Dy{\tilde{h}}^\hermitian \{\Dy{h}\{\X\}\} = \X$. It is not difficult to see that this equality is satisfied when
\begin{equation} \label{eq:duality_condition}
\sum_{z_\lambda,z_\omega} \tilde{h}_{z_\lambda,z_\omega}(\ll,\kk)\,\overline{h_{z_\lambda,z_\omega}}(\ll,\kk)=1, \quad \forall \: \ll, \kk.
\end{equation}
Although redundant joint time-vertex frames admit an infinite number of dual kernel sets satisfying \eqref{eq:duality_condition}, the typical choice is to use the \emph{canonical dual}, defined as 
\begin{equation} \label{eq:can_dual}
    \tilde{h}_{z_\lambda,z_\omega}(\ll,\kk) = \left( \sum_{z_{\lambda}',z_{\omega}'} h_{z_{\lambda}',z_{\omega}'}^2(\ll,\kk) \right)^{-1} \hspace{-3mm} h_{z_\lambda,z_\omega}(\ll,\kk).
\end{equation}
In fact, this corresponds to the pseudo-inverse of $\Dy{h}$, i.e., $\Dy{h}^\dagger = \Dy{\tilde{h}}^\hermitian$, while also having a low computational complexity.

To summarize, given an invertible time-vertex transform $\Dy{h}$ and coefficients $\C$, the inverse transform of $\Dy{h}$ associated with the set of kernels $\{h_{z_\lambda,z_\omega}\}_{[z_\lambda,z_\omega]\in\mathcal{Z}}$ is
\begin{equation}
\X =  \Dy{\tilde{h}}^{\hermitian}\left\{\C\right\}= \sum_{z_\lambda,z_\omega}\tilde{h}_{z_\lambda,z_\omega}(\LG,\LT)\C_{z_\lambda,z_\omega},
\end{equation}
where $\tilde{h}$ is defined in~\eqref{eq:can_dual}.

\section{Experiments}\label{sec:eval}

The suitability of the time-vertex framework for several classes of problems is illustrated on a wide variety of datasets: (a) dynamic meshes representing a walking dog and a dancing man, (b) the Caltrans Performance Measurement System (PeMS) traffic dataset depicting high resolution daily vehicle flow of 10 consecutive days in the highways of Sacramento measured every 5 minutes, (c) simulated SEIR- or SEIRS-type epidemics over Europe, (d) the Kuala Lumpur City Centre (KLCC) time-lapse video and (e) earthquake waveforms recorded by seismic stations geographically distributed in New Zealand, connected to the GeoNet Network.

Results suggest that joint analysis of time-vertex signals can bring forth benefits in signal denoising and recovery, learning and source localization problems. We remark that all the experiments were done using the GSPBOX~\cite{perraudin2014gspbox}, the UNLocBoX~\cite{perraudin2014unlocbox} and the LTFAT~\cite{prusa2014large}. Code reproducing the experiments is available at~\url{https://lts2.epfl.ch/reproducible-research/a-time-vertex-signal-processing-framework/}.

\begin{figure}
    \centering
    \includegraphics[width=0.5\textwidth]{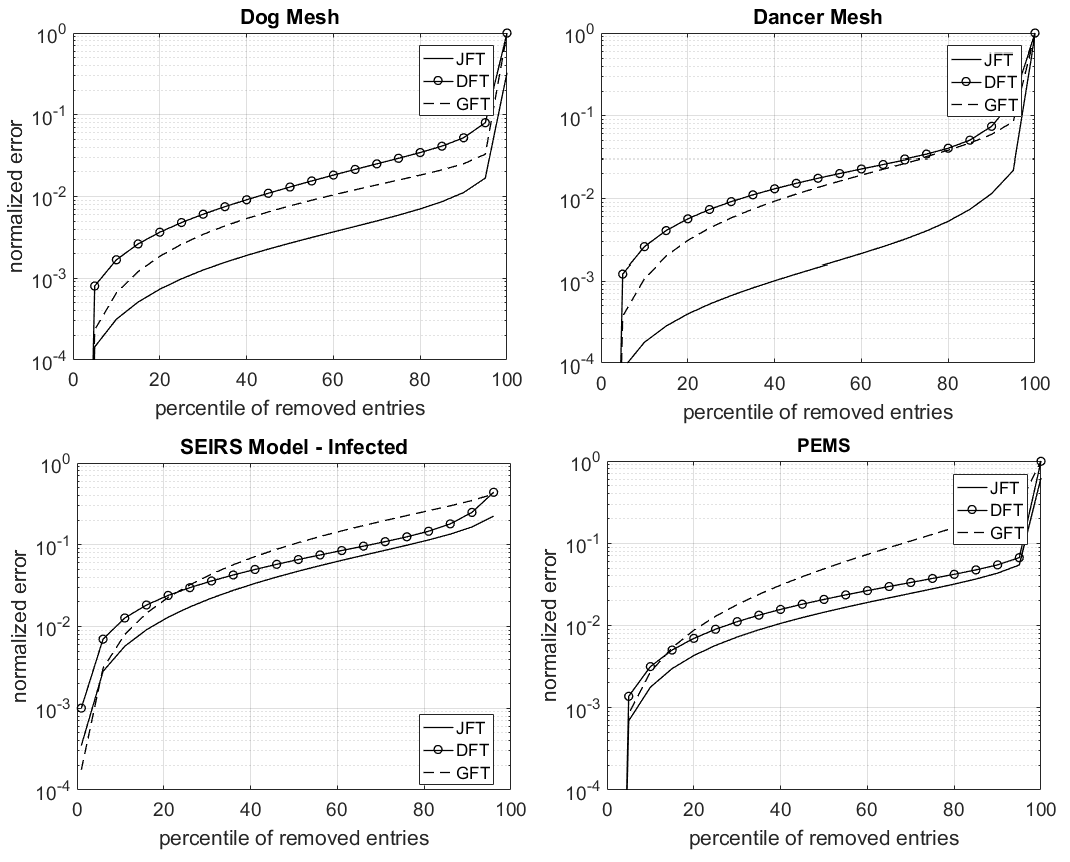}
    \caption{Compactness of the transforms for different datasets: dog and dancer meshes (above) number of infected over Europe according to SEIRS model (below left) and traffic flow measured by the PeMS (below right). Normalized error is computed reconstruting the signal after thresholding the values of the transforms below the $p$-th percentile.}
    \label{fig:sparsity}
\end{figure}

\subsection{Compactness of representation}
A key motivation behind the joint harmonic analysis is the capability of encoding time-varying graph-dependent signal evolution in a compact way. Our first step will therefore be to examine the energy compaction of the JFT transform in four datasets: two meshes representing a dancer ($N = 1502$ points in $\Rbb^3$ and $ T =  570$ timesteps) and a dog walking ($N = 2502$ points in $\Rbb^3$ and $ T =  52$ timesteps), the PeMS traffic flow dataset ($N = 710$ stations measuring traffic over $T = 2880$ intervals of 5-minute length each) and the number of infected individuals in an SEIRS epidemic (see Section~\ref{ssec:epidem} for description). Transforms with good energy compaction are desirable because they summarize the data well and can be used to construct efficient regularizers for regression problems.   

To measure energy compactness, we compute the DFT, GFT and JFT for each dataset, we replace the spectrum coefficients with magnitudes smaller than the $p$-th percentile with zeros and perform the corresponding inverse transform on the resulting coefficients. Denoting by $\X$ the original signal and $\X_p$ the compressed one, the compression error is for each $p$ given by $\norm{\X_p-\X}_F/\norm{\X}_F$. 
As shown in Figure~\ref{fig:sparsity}, JFT exhibits better energy compaction properties in all the datasets, and especially for the meshes where the graph captures well the signal structure.

\subsection{Regression problems with joint variation priors}

We next examine the utility of joint variation priors for regression problems in two example applications.

\paragraph{Denoising of dynamic meshes.}
\label{ssec:denoising}
Whenever a smoothness prior can be assumed, the joint Tikhonov regularization can be used to denoise a time-varying graph signal. The prior can be easily expressed in the time-vertex domain thanks to Eq.~\eqref{eq:tik}. Joint denoising is then performed by solving the following optimization problem
\begin{equation}\label{eq:denoising}
    \arg\min_{\X} \, \norm{\X-\Y}_{F}^{2}+\tau_1\norm{\GG \X}_F^2+\tau_2\norm{\X\GT}_F^2,
\end{equation}
where the regularization terms require the solution to be smooth in both graph and time domains. This problem has a closed form solution in our framework, which is a joint non-separable lowpass filter
\begin{equation}\label{eq:tikfilter}
    h_{\mathrm{TIK}}(\ll, \kk) = \dfrac{1}{1+\tau_1 \ll + 2\tau_2 (1-\cos(2\pi\kk))}.
\end{equation}
\begin{figure}
   \includegraphics[width=\columnwidth]{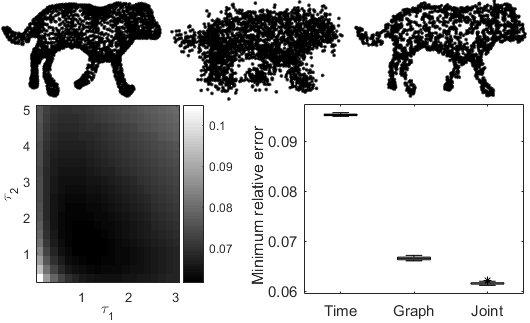}
    \caption{Joint variation priors are useful in denoising the coordinates of dynamic meshes. The original mesh (left) was corrupted with random Gaussian noise (center, normalized error 0.2); shown here for one realization of the noise. After the denoising, the error decreases to 0.06 (right). The normalized error as a function of parameters $\tau_1$ and $\tau_2$ is shown in a heat-map (below left) averaged over 20 realizations. The boxplot (below right) shows the minimum achievable error for a time ($\tau=0$), graph ($\tau_2 = 0$) and joint variation prior.}
    \label{fig:dog_denoising}
\end{figure}
In order to investigate its performance, we consider the vertices of a mesh of size ${2502 \times 59 \times 3}$ representing a walking dog, we add Gaussian noise to the coordinates, we build a $k$ nearest neighbor graph based on the distances between the time average of the coordinates and finally we solve the problem~\eqref{eq:denoising} for each coordinate dimension. We averaged the results over 20 realizations of the noise. 

The meshes in Figure~\ref{fig:dog_denoising} represent, from left to right, the original, the noisy and the recovered one, for one realization of the noise. Remarkably, the normalized error drops from 0.20 to 0.06, respectively before and after denoising, making the dog distinguishable again. As side effect, the dog appears to be thinner, due to the graph regularization. The heat-map in the left corner of the figure shows the role of the regularization parameters. We found (using exhaustive search) that the lowest error is achieved when  $\tau_1 = 0.71$ and $\tau_2 = 1.78$. We compare the performance of the joint Tikhonov regularization with respect to time- and graph-only for the best parameter combinations of all methods. The boxplot on the right shows the minimum achievable error statistics in the three cases over the 20 realizations. It is easy to see that the graph plays a major role in the denoising, since it encodes the structural information of the mesh. Nevertheless, the joint approach performs the best, i.e., $0.062\pm 0.0002$, taking advantage of the smoothness in both domains, while graph and time methods achieve $0.067\pm 0.0003$ and  $0.095\pm 0.0002$, respectively.

\begin{figure}
    \centering
    \includegraphics[width=\columnwidth]{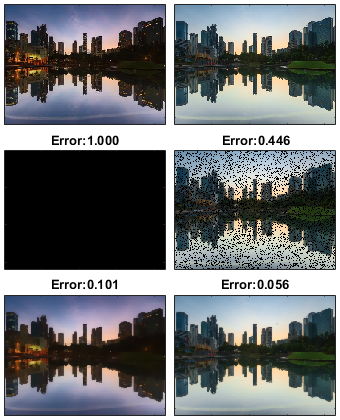}
    \caption{Visual inspection of video inpainting results. We show two frames extracted from the video (top), their corrupted counterparts (middle left frame is missing entirely, whereas middle right frame has missing pixels), and the reconstructions using our method with an $N_{1,2}$ regularizer (bottom).}
    \label{fig:video}
\end{figure}

\paragraph{Inpainting of time-lapse video.}
\label{ssec:inpainting}
We consider the problem of time-vertex signal recovery from noisy, corrupted, and incomplete measurements. Depending on the characteristics of the signal, the prior $\jnorm{\x}{p}{q}$ with different values of $p$ and $q$ and different weights can be used. A typical signal recovery problem in signal processing is the image inpainting, i.e., trying to replace corrupted or lost part of the image. Since patch-graphs allow non-local image processing \cite{buades2005nonlocal}, our goal is to extend graph-based non-local processing to video inpainting and recovery. However, since our framework is constrained to static graphs, we focus on the particular case of time-lapse videos, whose structure stays majorly invariant throughout the video. 
To this end, we corrupted a time-lapse video that shows the skyline of the Kuala Lumpur City Centre, which statistical properties were amenable from a graph perspective, being the skyline static with time-varying colors. The video has size ${160\times 214\times 3 \times 604}$ (height $\times$ width $\times$ colors $\times$ frames). We removed $20\%$ of the pixels and $20\%$ of the frames from the original KLCC video, achieving a normalized error of $0.61$. 
The inpainting is performed solving the optimization problem for each color using as regularizer $\jnorm{\x}{1}{2}$: 
\begin{equation}\label{eq:inpainting}
    \arg\min_{\X} \, \norm{\M\circ \X-\Y}_{F}^{2}+\gamma_1\norm{\GG \X}_1+\gamma_2\norm{\X\GT}_F^2,
\end{equation}
where $\M$ is the mask of the missing entries. The patch graph $G$ is constructed from the video averaged in time. The rationale is that the $l_1$ over the graph will restore the missing pixels, being each frame approximately piece-wise constant, whereas the $l_2$ norm in time recovers the smooth changes of the colors from dawn until dusk. 

\begin{figure}
\centering
    \includegraphics[width=\columnwidth]{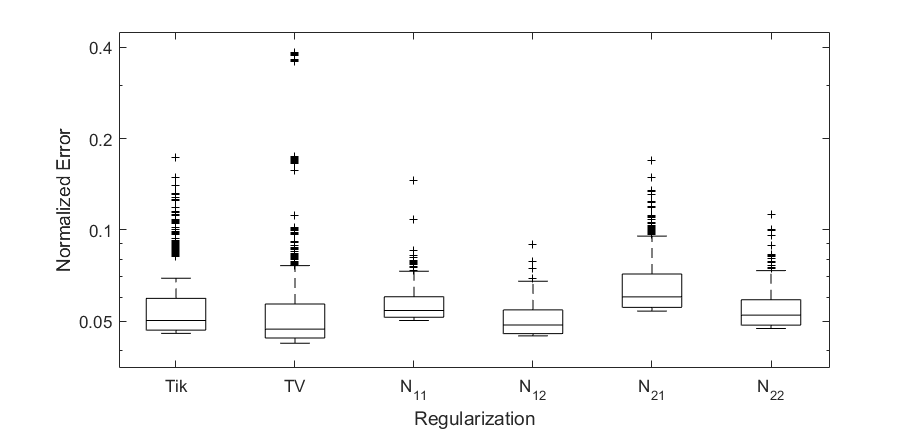}
    \caption{Comparison of video inpainting performances between Tikhonov, TV and Joint regularizations. Each box represents a statistical summary of the error evaluated for each frame of the whole video. Although TV achieves the best recovery for some frames, in case of occluded frames the error is very large. Joint regularization $\jnorm{\x}{1}{2}$ trades the lowest error achievable with a better average recovery.}
    \label{fig:video_comparison}
\end{figure}
\newcolumntype{L}[1]{>{\raggedright\arraybackslash}p{#1}}
\newcolumntype{C}[1]{>{\centering\arraybackslash}p{#1}}
\newcolumntype{R}[1]{>{\raggedleft\arraybackslash}p{#1}}
\begin{table}[h]
\renewcommand{\arraystretch}{1.3}
 \caption{Video inpainting normalized errors}
    \label{tab:video_comparison}
    \centering
    \begin{tabular}{L{2cm} C{1.5cm} C{1.5cm} C{0.7cm}}
        \hline
        Regularizer  & Pixels & Frames & Total\\
        \hline
        Tikhonov  & 0.051          & 0.100          & 0.059  \\
        TV        & \textbf{0.048} & 0.122          & 0.060  \\
        $\jnorm{\x}{1}{1}$     & 0.056 &          0.059 & 0.057 \\
        $\jnorm{\x}{1}{2}$     & 0.050 & \textbf{0.055} & \textbf{0.051}  \\
        $\jnorm{\x}{2}{1}$     & 0.061 & 0.103  & 0.068 \\
        $\jnorm{\x}{2}{2}$     & 0.053 & 0.066  & 0.055 
\end{tabular}
\end{table}

Figure~\ref{fig:video} shows two frames of the video, their corrupted counterparts and the result of the recovery, along with the respective normalized errors. The recovered video has a normalized error of $0.049$, illustrating that the joint inpainting is able to restore the global quality of the video even in case of considerable missing information.

We compare the recovery performance with all the joint regularizers $\jnorm{\x}{p}{q}$ for $p,q = \{1,2\}$, and with two baseline algorithms, based on 3D-Tikhonov and isotropic 3D-TV regularizations~\cite{chan2011augmented}. The last two correspond to using a grid graph with equal weights on the edges. Table~\ref{tab:video_comparison} reports the normalized errors averaged over the pixels-only, frames-only and the whole video. The better performance achieved by the joint regularizer $\jnorm{\x}{1}{2}$ is due to its capability to restore missing frames, while missing pixels recovery performances are almost the same. Figure~\ref{fig:video_comparison} illustrates a summary statistics of the errors computed over each frame. Although TV performs the best in the median, in case of occluded frames the error is much larger w.r.t. the joint recovery, leading to a higher average error.

\subsection{Overcomplete representations}

Last, we examine the utility of STVFT and STVWT, respectively, as a feature extractor for dynamic mesh clustering and as a dictionary used to uncover the wave-like structure and epicenter of a seismic event.

\begin{figure}[t!]
    \centering
    \includegraphics[width=\columnwidth]{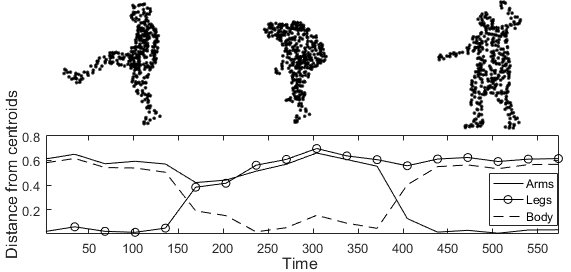}
    \caption{Clustering of the dancer mesh (no noise): the plot (below) shows the distance of the points stemming from the STVFT representation of three frames (above) of the time-varying mesh closest to the clusters centroids. Each frame shows a different phase of the dance.}
    \label{fig:dancer_cluster}
\end{figure}

\paragraph{Clustering dynamic meshes using STVFT.}
\label{ssec:clustering}
We consider the motion classification of a dynamic mesh representing the dancer, corrupted with additive sparse noise with density 0.1 with normally distributed entries and SNR of $-20$ dB and $-10$ dB. Our objective is to determine the phase of the dance (\textit{moving arms}, \textit{stretching legs} and \textit{bending body}) at each frames by performing spectral clustering on some representation of the windowed signal. To obtain the ground-truth, we labeled each frame by hand and verified that, when the noiseless signal (i.e., the actual trajectory of the points in time over each window) was used to define the features, one obtains a classification accuracy of 0.926. 

Since we want to localize spatial-structured phenomena in time, our approach will be to use a STVFT to derive the representation. 
To capture the geometry of the problem, we used a nearest neighbour graph constructed based on the coordinates of the mesh vertices averaged in time; this graph was fixed for the whole sequence. 
As explained in Section~\ref{sec:JTVD}, the STVFT is separable, meaning that we can handle the vertex and the time dimensions separately. In the time domain we use a rectangular window with support equal to $50$ samples in time and spacing such that the overlap is $60\%$.
For the vertex dimension, we use the an Itersine kernel (defined in the GSPBOX~\cite{perraudin2014gspbox}) that we uniformly translate at 5 different positions in the graph spectral domain. 

The STVFT provides features associated to a time instant that we can directly use to classify the dance (see Figure~\ref{fig:dancer_cluster} for a visual illustration of the clustering results). Other transforms such as GFT, DFT, and JFT do no have this property. Hence, in order to compare with these other transforms, we use the same rectangular windows (width 50, overlap 20 samples) to extract 27 time sequences from the signal. We then used the transformed data associated with each sequence as a point to be clustered.
\begin{figure}[t!]
    \centering
    \includegraphics[width=0.95\columnwidth]{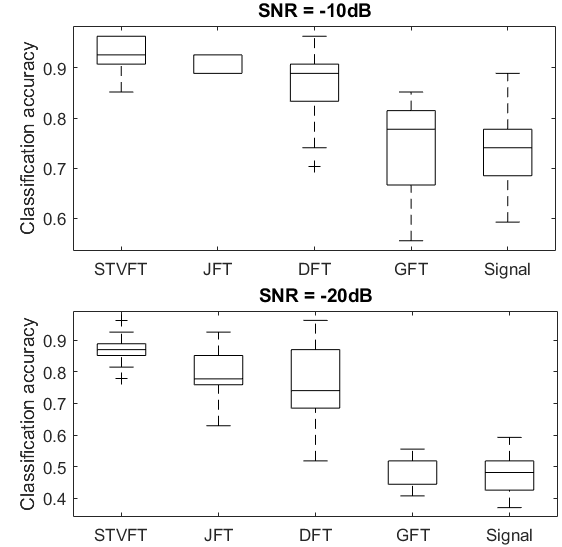}
    \caption{Comparison of clustering accuracy using different transforms in case of sparse Gaussian noise for SNR -20 dB and -10 dB. Each box shows a summary statistics of the accuracy computed over 20 different realizations of the noise. Results show that SVTFT achieves the highest accuracy in average.}
    \label{fig:dancer_cluster_accuracy}
\vspace{-0.5cm}
\end{figure}
\begin{figure*}[t]
\centering
    \includegraphics[width=0.3\textwidth]{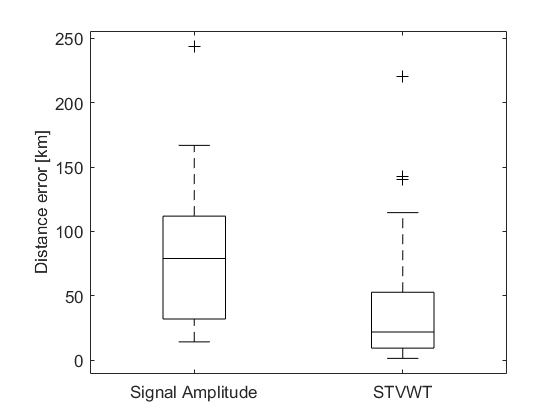} 
    \includegraphics[width=0.69\textwidth]{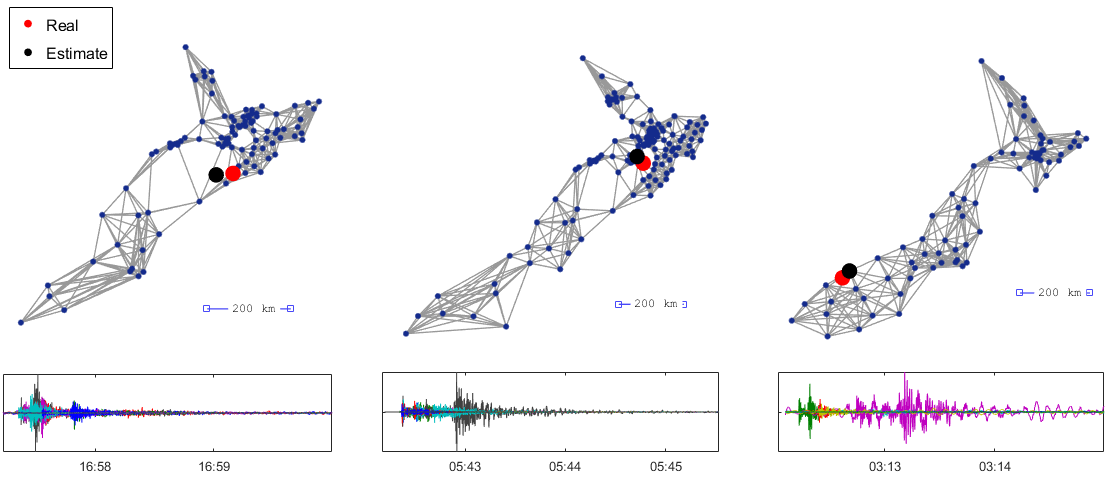}
    \caption{Left: Comparison of seismic epicenter localization performances between amplitude-based approach and STVWT. The bar graph shows that the second outperforms the first, suggesting that the damped wave model assumption significantly improves the source estimation performance. 
    Right: Results for 3 different seismic events in New Zealand. Right top: the graph is created using the coordinates of the available stations for each event and connecting the closest stations. The stars and the circles are the true and estimated sources of the seismic wave respectively. Right bottom:  Signal recorded by the sensors over time for each event.}\label{fig:seismic}
\end{figure*}

Figure~\ref{fig:dancer_cluster_accuracy} illustrates the clustering accuracy statistics over 20 realizations of sparse noise for features constructed based on the magnitude of five representations: the windowed sequences, as well as their GFT, DFT, JFT, and STVFT representations.
Observe how the presence of sparse noise severely hampers classification when the raw signal is used, with the average accuracy dropping from 0.926 for the clean signals to 0.469 and 0.74 for $-20$ dB and $-10$ dB, respectively. 

We can also see that the two representations leading to the highest median accuracy are the JFT and STVFT, suggesting the utility of joint harmonic representations. Nevertheless, the STVFT provides more robust estimates with an average accuracy of 0.869 rather than 0.792 for the JFT at $-20$ dB.

\paragraph{Seismic epicenter estimation with STVWT.}
\label{ssec:source}
We analyze seismic events recorded by the GeoNet sensor network whose epicenters were chosen to be randomly distributed in different areas of New Zealand. We extend the results presented in~\cite{grassi2016tracking} to a greater dataset using the STVWT with mother kernel based on the wave PDE, which allows us to decompose the signal as sum of PDE solutions. As a first approximation, when the waves propagate in a continuous domain or a regular lattice, seismic waveforms can be modeled as oscillating damped waves~\cite{lowrie2007fundamentals}. Our premise is thus that we can approximate the seismic waveforms using a small set of damped waves propagating on the graph connecting the seismic stations. Thus, we expect the damped wave mother kernel 
\begin{equation}
 h(\ll,\kk) =\frac{1}{\sqrt{T}}\frac{\eu^{\beta+\ju\kk}+ \ll/2-1}{2 (\cosh\lrpar{\beta+\ju\kk}+ \ll/2-1)} 
\end{equation} 
to be a good approximation of the seismic waves recorded by the sensors, with the damping factor $\beta$ chosen to fit the damping present in the seismic signals. To construct the STVWT, we select 10 equally spaced values in $[0,2]$ for $z_\lambda$ and set $z_\omega=1$.
To estimate the epicenter of the earthquake we solve 
\begin{equation} \label{eq:opt_l1_dict}
    \arg\min_{\C}\|\Dy{h}^{\hermitian}\{\C\}-\X\|_{2}^{2}+\gamma\|\C\|_{1},
\end{equation}
where $\gamma$ is the regularization parameter controlling the trade-off between the fidelity term (selected using exhaustive search) and the sparseness assumption and $\Dy{h}^{\hermitian}$ is the synthesis operator associated with STVWT. 

The solution provides important pieces of information. Firstly, using the synthesis operator we can obtain a denoised version of the original process. Secondly, the non-zero coefficients of $\C$, describe the origins and amplitudes of the different components. Therefore, we average the coordinates of the vertices corresponding to the sources of the waves with highest energy coefficients. We compare the performance of STVWT with the estimate obtained using only the signal amplitude: for each earthquake we average the coordinates of the stations using as weights  energy of the signals. Figure~\ref{fig:seismic} shows on the left the comparison over 40 different seismic events randomly distributed over the New Zealand between the two methods. STVWT based on the damped wave kernel achieves an average error of 48.5 km, providing an almost twofold improvement over the baseline, whose average performance is 88.3 km. On the right, it illustrates the estimate for 3 different seismic events and the respective seismic waveforms. These results show that the proposed method significantly improves the source estimation performance.


\section{Conclusion}
This work puts forth a Time-Vertex Signal Processing Framework, that facilitates the analysis of graph structured data that also evolve in time. 
We motivate our framework leveraging the notion of partial differential equation on graphs. We introduce joint operators, such as time-vertex localization and we present a novel approach to significantly improve the accuracy of fast joint filtering. We also illustrate how to build time-vertex dictionaries, providing conditions for efficient invertibility and examples of constructions. 
Our experimental results on a variety of datasets, suggest that the proposed tools can bring forth significant benefits in various signal processing and learning tasks involving time-series on graphs.

\appendix

\section{Appendices}
\label{ssec:proof}

\subsection{Wave equation}

In the continuous setting, the wave equation is
\begin{equation*}
    \partial_{tt} u - \Delta u = 0    
\end{equation*}
where $ {u \colon \Rbb \times \Rbb^d \rightarrow \Cbb}$ is a function of both time ${t \in \Rbb}$ and space ${x \in \Rbb^d}$, with ${\Delta}$ being the Laplacian operator. 
Even though being a second order PDE, the equation can be rewritten as a first order system defining  ${v(t) := \partial_t u(t)}$. The pair ${(u(t),v(t))}$ evolves now according to the system
\begin{equation}
\left\lbrace
\begin{array}{l}
\partial_t u(t) = v(t)\\[1em]
\partial_t v(t) = \Delta u(t)
\end{array}
\right.
\end{equation}
Assuming vanishing initial velocity ${v(0)=0}$, the solution ${u(t)}$ is given via functional calculus by~\cite{durran2013numerical}
\begin{equation}\label{eq:cwave_sol}
    u(t) = \cos(t \sqrt{-\Delta}) u(0)
\end{equation}
where $\cos(t \sqrt{-\Delta})$ is called \emph{propagator operator}.

To obtain a discrete wave equation evolving on a graph, we approximate the second order time derivative with its stencil approximation and the continuous Laplacian $\Delta$ with the graph Laplacian $\LG$ with reversed sign:
%
%
\begin{equation}\label{eq:wavepde}
\X \LT=s \LG \X,
\end{equation}
where $s>0$ is the speed of the propagation.
The wave equation is a hyperbolic differential equation and several difficulties arise when discretizing it for numerical computation of the solution\cite{durran2013numerical}. Moreover, the graph being an irregular domain, the solution of the above equation is not  any more a smooth wave after a few iterations.
Nevertheless, we assume as in the case of the heat diffusion Eq.~\eqref{eq:heat} that the solution can be written as
\begin{equation}\label{eq:wave_sol}
    \x_t=K_{s}(\LG,t)\x_1=\K_{t,s}\x_1,
\end{equation}
where ${\K_{t,s}=K_{s}(\LG,t)}$ is a matrix obtained applying the function $K_{s}(\LG,t)$ to the scaled Laplacian $s\LG$ and parametrized by the time $t$. We will call the operator $\K_{t,s}$ ``the discrete analogue of the wave propagator'' of Eq.~\eqref{eq:cwave_sol}. 
Therefore, matrix ${\X=\lrsquare{\K_{t,s}\x_1}_{t=1}^T=\Ker{s}{\x_1}}$ is obtained stacking the vectors $\x_t$ of Eq.~\eqref{eq:wave_sol} along the columns. 
Substituting \eqref{eq:wave_sol} into \eqref{eq:wavepde}, we obtain $\Ker{s}{\x_1}\L_T=s \LG \Ker{s}{\x_1}$
%
%
which in the graph spectral domain is
\begin{equation} \label{eq:condpde}
    \GKer{s}{\tilde{\x}_1} \LT  = s\b{\Lambda}_G \GKer{s}{\tilde{\x}_1},
\end{equation}
where ${\widetilde{\K}_{t,s}=K_{s}(\b{\Lambda}_G,t)}$. Equation~\eqref{eq:condpde} is formally analogous to the eigendecomposition of the operator $\LT$, therefore, the $\ell$-th row of $\GKer{s}{\tilde{\x}_1}$ must be an eigenvector of $\L_T$ with eigenvalue $\ll$, for every $\ell$.
It has been proved in~\cite{strang1999discrete} that the eigenvectors ${\UT(t,k) = \cos(t k\pi/T)}$ form also the discrete Fourier basis. Using Eq.~\eqref{eq:dctfreq}, we obtain
\begin{equation}\label{eq:wave_ker2}
K_{s}(\ll,t) = 
\cos(t\tl),
\end{equation}
with $\tl = \arccos(1-\frac{s\ll}{2})$.
Since the $\arccos(x)$ is defined only for ${x\in\lrsquare{-1,1}}$, to guarantee stability the parameter $s$ must satisfy ${s<4/\lambda_{max}}$. We remark that this result is in agreement with the stability analysis of numerical solver for the discrete wave equation presented in~\cite{durran2013numerical}. 

Taking the DFT of the wave kernel in Eq.~\eqref{eq:wave_ker2}, we obtain
\begin{equation*}
\widehat{K}_{s}(\ll,\kk) =	\sum_{t}\cos(t\tl)\eu^{-\ju \kk t}
\end{equation*}
Therefore, the solution in the joint spectral domain can be written as
\begin{equation*}
    \Xhat(\ell,k) = \widehat{K}_{s}(\ll,\kk)\Z(\ell,k),
\end{equation*}
where ${\Z({\ell,k}) = \widetilde{\x_1}(\ell) \, \U_T^*(k,1)}$.

Note that there exists a closed form solution for the function $\widehat{K}_{s}$:
\begin{align*}
   &\widehat{K}_{s}(\ll,\kk) = \\
  & \begin{cases}
  \dfrac{\delta\lrpar{\kk + \tl} +\delta\lrpar{\kk - \tl}}{2}, \hspace{4em} \text{if } T\tl/2\pi \text{ integer} &\\[1em]
    \dfrac{1}{2}\lrpar{\dfrac{1-\eu^{-\ju T(\kk+\tl)}}{1-\eu^{-\ju(\kk+\tl)}} + \dfrac{1-\eu^{-\ju T(\kk-\tl)}}{1-\eu^{-\ju(\kk-\tl)}}}, \, \text{ otherwise} &
  \end{cases} 
\end{align*}
\subsection{Frame bound for joint time-vertex dictionaries}

\begingroup
\def\thetheorem{\ref{theo:frame}}
\begin{theorem}
Let $\lrbrace{h_z(\lambda, \omega)}_{z \in \mathcal{Z}_\lambda \times \mathcal{Z}_\omega}$ be the kernels of a time-vertex dictionary $\mathcal{D}_{h}$, and set
\begin{align*}
A=\min_{l,k}\sum_{z}\abs{h_{z}(\ll,\omega_{k})}^{2}, \
B=\max_{l,k}\sum_{z}\abs{h_{z}(\ll,\omega_{k})}^{2}.
\end{align*} 
If $0 < A \leq B < \infty$, then $\Dy{h}$ is a frame in the sense:
\begin{equation*}
A\norm{\X}_F^2\leq \norm*{\Dy{h}\{\X\}}_F^2 \leq B\norm{\X}_F^2
\end{equation*}
for any time-vertex signal $\X\in\Rbb^{N\times T}$.
\end{theorem}
\addtocounter{theorem}{-1}
\endgroup
\begin{proof}
In the joint spectral domain we can write:
\begin{align*}
 &\norm*{\left\{\Dy{h}\{\X\}\right\}}^F_2 = \sum_{m, \tau,z} \abs*{\left\{ \Dy{h}\{\X\}\right\} (m,\tau,z)}^2\\
 &=\sum_{z, m, \tau}\lrpar{\sum_{\substack{\ell,k\\n,t}} \X(n,t) h_{z}(\ll,\kk) \u_{\ell}^{\hermitian}(n)\u_{\ell}(m)\eu^{-\ju\kk({t-\tau})}}\\
 &\lrpar{\sum_{\substack{\ell',k'\\n',t'}} \X(n',t')  h_{z}(\lambda_{\ell'},\omega_{k'}) \u_{\ell'}^{\hermitian}(n')\u_{\ell'}(m)\eu^{-\ju\omega_{k'}({t'-\tau})}}^{\hermitian}\\
 &=\sum_{z,\ell, k} h_{z}(\ll,\kk) \widehat{h}_{z}^{\hermitian}(\ll,\kk) \widehat{\X}(\ell, k)\widehat{\X}^{\hermitian}(\ell, k)\\
 &=\sum_{z,\ell, k} \abs{h_{z}(\ll,\kk)}^{2}\abs{\widehat{\X}(\ell, k)}^{2}=\sum_{z}\langle\abs{h_{z}}^2, \abs{\widehat{\X}}^2\rangle,
 \end{align*}
where the equality holds due to the orthogonality of the eigenvectors. Finally, each element in the sum can be lower bounded and upper bounded by the minimum and maximum value that every filter takes over $\ell$ and $k$. Using the Parseval relation~\eqref{eq:parseval}, the theorem holds.
\end{proof}

\bibliographystyle{IEEEtran}
\bibliography{bibliography}

\begin{thebibliography}{10}
\providecommand{\url}[1]{#1}
\csname url@samestyle\endcsname
\providecommand{\newblock}{\relax}
\providecommand{\bibinfo}[2]{#2}
\providecommand{\BIBentrySTDinterwordspacing}{\spaceskip=0pt\relax}
\providecommand{\BIBentryALTinterwordstretchfactor}{4}
\providecommand{\BIBentryALTinterwordspacing}{\spaceskip=\fontdimen2\font plus
\BIBentryALTinterwordstretchfactor\fontdimen3\font minus
  \fontdimen4\font\relax}
\providecommand{\BIBforeignlanguage}[2]{{%
\expandafter\ifx\csname l@#1\endcsname\relax
\typeout{** WARNING: IEEEtran.bst: No hyphenation pattern has been}%
\typeout{** loaded for the language `#1'. Using the pattern for}%
\typeout{** the default language instead.}%
\else
\language=\csname l@#1\endcsname
\fi
#2}}
\providecommand{\BIBdecl}{\relax}
\BIBdecl

\bibitem{de2013anatomy}
M.~{De Domenico}, A.~Lima, P.~Mougel, and M.~Musolesi, ``{The anatomy of a
  scientific rumor},'' \emph{Scientific reports}, 2013.

\bibitem{adamic2005political}
L.~A. Adamic and N.~Glance, ``{The political blogosphere and the 2004 US
  election: divided they blog},'' in \emph{Proceedings of the 3rd international
  workshop on Link discovery}.\hskip 1em plus 0.5em minus 0.4em\relax ACM,
  2005, pp. 36--43.

\bibitem{guille2013information}
A.~Guille, H.~Hacid, C.~Favre, and D.~A. Zighed, ``{Information diffusion in
  online social networks: A survey},'' \emph{ACM SIGMOD Record}, no.~2, pp.
  17--28, 2013.

\bibitem{mohan2008nericell}
P.~Mohan, V.~N. Padmanabhan, and R.~Ramjee, ``Nericell: rich monitoring of road
  and traffic conditions using mobile smartphones,'' in \emph{SENSYS}.\hskip
  1em plus 0.5em minus 0.4em\relax ACM, 2008, pp. 323--336.

\bibitem{RevModPhys.87.925}
R.~Pastor-Satorras, C.~Castellano, P.~{Van Mieghem}, and A.~Vespignani,
  ``{Epidemic processes in complex networks},'' \emph{Rev. Mod. Phys.}, no.~3,
  pp. 925--979, aug 2015.

\bibitem{huang2015graph}
W.~Huang, L.~Goldsberry, N.~F. Wymbs, S.~T. Grafton, D.~S. Bassett, and
  A.~Ribeiro, ``Graph frequency analysis of brain signals,'' \emph{IEEE Journal
  of Selected Topics in Signal Processing}, no.~7, pp. 1189--1203, 2016.

\bibitem{smith2016physiological}
K.~Smith, B.~Ricaud, N.~Shahid, S.~Rhodes, J.~M. Starr, A.~Ibanez, M.~A. Parra,
  J.~Escudero, and P.~Vandergheynst, ``The physiological underpinnings of
  visual short-term memory binding using graph modular dirichlet energy:
  Evidence from healthy subjects,'' \emph{arXiv preprint arXiv:1606.02587},
  2016.

\bibitem{shuman2013emerging}
D.~I. Shuman, S.~K. Narang, P.~Frossard, A.~Ortega, and P.~Vandergheynst,
  ``{The emerging field of signal processing on graphs: Extending
  high-dimensional data analysis to networks and other irregular domains},''
  \emph{Signal Process. Mag., IEEE}, no.~3, pp. 83--98, 2013.

\bibitem{sandryhaila2013discrete}
A.~Sandryhaila and J.~M.~F. Moura, ``{Discrete signal processing on graphs},''
  \emph{IEEE Trans. Signal Process.}, no.~7, pp. 1644--1656, 2013.

\bibitem{belkin2001laplacian}
M.~Belkin and P.~Niyogi, ``Laplacian eigenmaps and spectral techniques for
  embedding and clustering.'' in \emph{NIPS}, 2001, pp. 585--591.

\bibitem{shahid2016fast}
N.~Shahid, N.~Perraudin, V.~Kalofolias, G.~Puy, and P.~Vandergheynst, ``Fast
  robust pca on graphs,'' \emph{IEEE Journal of Selected Topics in Signal
  Processing}, no.~4, pp. 740--756, 2016.

\bibitem{perraudin2016stationary}
N.~Perraudin and P.~Vandergheynst, ``{Stationary signal processing on
  graphs},'' \emph{IEEE Trans. Signal Process.}, no.~99, pp. 1--1, 2017.

\bibitem{marques2016stationary}
A.~G. Marques, S.~Segarra, G.~Leus, and A.~Ribeiro, ``Stationary graph
  processes and spectral estimation,'' \emph{arXiv preprint arXiv:1603.04667},
  2016.

\bibitem{hammond2011wavelets}
D.~K. Hammond, P.~Vandergheynst, and R.~Gribonval, ``{Wavelets on graphs via
  spectral graph theory},'' \emph{Applied and Computational Harmonic Analysis},
  no.~2, pp. 129--150, 2011.

\bibitem{coifman2006diffusion}
R.~R. Coifman and M.~Maggioni, ``{Diffusion wavelets},'' \emph{Applied and
  Computational Harmonic Analysis}, no.~1, pp. 53--94, 2006.

\bibitem{belkin2004semi}
M.~Belkin and P.~Niyogi, ``{Semi-supervised learning on Riemannian
  manifolds},'' \emph{Machine learning}, no. 1-3, pp. 209--239, 2004.

\bibitem{smola2003kernels}
A.~J. Smola and R.~Kondor, ``{Kernels and regularization on graphs},'' in
  \emph{Learning theory and kernel machines}, B.~Sch{\"{o}}lkopf and
  M.~Warmuth, Eds.\hskip 1em plus 0.5em minus 0.4em\relax Springer, 2003, pp.
  144--158.

\bibitem{kalofolias2016learn}
V.~Kalofolias, ``How to learn a graph from smooth signals,'' in \emph{AISTATS},
  2016.

\bibitem{loukas2016frequency}
A.~Loukas and D.~Foucard, ``{Frequency Analysis of Temporal Graph Signals},''
  in \emph{GLOBALSIP}, 2016.

\bibitem{isufi2016separable}
E.~Isufi, A.~Loukas, A.~Simonetto, and G.~Leus, ``Separable autoregressive
  moving average graph-temporal filters,'' in \emph{EUSIPCO}.\hskip 1em plus
  0.5em minus 0.4em\relax IEEE, 2016, pp. 200--204.

\bibitem{grassi2016tracking}
F.~Grassi, N.~Perraudin, and B.~Ricaud, ``Tracking time-vertex propagation
  using dynamic graph wavelets,'' in \emph{GLOBALSIP}, 2016.

\bibitem{dahlhaus2003causality}
R.~Dahlhaus and M.~Eichler, ``Causality and graphical models in time series
  analysis,'' \emph{Oxford Statistical Science Series}, pp. 115--137, 2003.

\bibitem{zhang2015graph}
C.~Zhang, D.~Flor{\^{e}}ncio, and P.~A. Chou, ``{Graph Signal Processing--A
  Probabilistic Framework},'' Microsoft Research Lab - Redmond, Tech. Rep.,
  2015.

\bibitem{sandryhaila2014big}
A.~Sandryhaila and J.~M.~F. Moura, ``{Big data analysis with signal processing
  on graphs: Representation and processing of massive data sets with irregular
  structure},'' \emph{IEEE Signal Process. Mag.}, no.~5, pp. 80--90, 2014.

\bibitem{kivela2014multilayer}
M.~Kivel{\"{a}}, A.~Arenas, M.~Barthelemy, J.~P. Gleeson, Y.~Moreno, and M.~A.
  Porter, ``{Multilayer networks},'' \emph{Journal of Complex Networks}, no.~3,
  pp. 203--271, 2014.

\bibitem{benzi2016principal}
K.~Benzi, B.~Ricaud, and P.~Vandergheynst, ``Principal patterns on graphs:
  Discovering coherent structures in datasets,'' \emph{IEEE Transactions on
  Signal and Information Processing over Networks}, no.~2, pp. 160--173, 2016.

\bibitem{de2013mathematical}
M.~{De Domenico}, A.~Sol{\'{e}}-Ribalta, E.~Cozzo, M.~Kivel{\"{a}}, Y.~Moreno,
  M.~A. Porter, S.~G{\'{o}}mez, and A.~Arenas, ``{Mathematical formulation of
  multilayer networks},'' \emph{Physical Review X}, no.~4, p. 41022, 2013.

\bibitem{loukas2015distributed}
A.~Loukas, A.~Simonetto, and G.~Leus, ``Distributed autoregressive moving
  average graph filters,'' \emph{IEEE Signal Process. Lett.}, no.~11, pp.
  1931--1935, 2015.

\bibitem{valdivia2015wavelet}
P.~Valdivia, F.~Dias, F.~Petronetto, C.~T. Silva, and L.~G. Nonato,
  ``Wavelet-based visualization of time-varying data on graphs,'' in \emph{2015
  IEEE Conference on Visual Analytics Science and Technology (VAST)}, Oct 2015,
  pp. 1--8.

\bibitem{mei2015signal}
J.~Mei and J.~M.~F. Moura, ``{Signal processing on graphs: Estimating the
  structure of a graph},'' in \emph{in ICASSP}.\hskip 1em plus 0.5em minus
  0.4em\relax IEEE, 2015, pp. 5495--5499.

\bibitem{isufi2017autoregressive}
E.~Isufi, A.~Loukas, A.~Simonetto, and G.~Leus, ``Autoregressive moving average
  graph filtering,'' \emph{IEEE Trans. Signal Process.}, no.~2, pp. 274--288,
  2017.

\bibitem{perraudin2016towards}
N.~Perraudin, A.~Loukas, F.~Grassi, and P.~Vandergheynst, ``Towards stationary
  time-vertex signal processing,'' in \emph{ICASSP}, 2017.

\bibitem{loukas2016predicting}
A.~Loukas and N.~Perraudin, ``Predicting the evolution of stationary graph
  signals,'' \emph{arXiv preprint arXiv:1607.03313}, 2016.

\bibitem{loukas2016stationary}
A.~{Loukas} and N.~{Perraudin}, ``{Stationary time-vertex signal processing},''
  \emph{ArXiv e-prints}, Nov. 2016.

\bibitem{burago2013graph}
D.~Burago, S.~Ivanov, and Y.~Kurylev, ``A graph discretization of the
  laplace-beltrami operator,'' \emph{Journal of Spectral Theory}, vol.~4,
  no.~4, pp. 675--714, 2014.

\bibitem{shuman2013vertex}
D.~I. Shuman, B.~Ricaud, and P.~Vandergheynst, ``{Vertex-frequency analysis on
  graphs},'' \emph{Applied and Computational Harmonic Analysis}, no.~2, pp.
  260--291, 2013.

\bibitem{merris1994laplacian}
R.~Merris, ``Laplacian matrices of graphs: a survey,'' \emph{Linear Algebra and
  its Applications}, pp. 143 -- 176, 1994.

\bibitem{solomon2015pde}
\BIBentryALTinterwordspacing
J.~Solomon, ``{PDE} approaches to graph analysis,'' \emph{CoRR}, 2015.
  [Online]. Available: \url{http://arxiv.org/abs/1505.00185}
\BIBentrySTDinterwordspacing

\bibitem{courant1967partial}
R.~Courant, K.~Friedrichs, and H.~Lewy, ``On the partial difference equations
  of mathematical physics,'' \emph{IBM Journal of Research and Development},
  no.~2, pp. 215--234, March 1967.

\bibitem{fourier1807memoire}
J.-B.-J. Fourier, ``M{\'e}moire sur la propagation de la chaleur dans les corps
  solides,'' \emph{Nouveau Bulletin des sciences par la Société philomatique
  de Paris}, p. 112–116, 1807.

\bibitem{narasimhan1999fourier}
\BIBentryALTinterwordspacing
T.~N. Narasimhan, ``Fourier's heat conduction equation: History, influence, and
  connections,'' \emph{Reviews of Geophysics}, no.~1, pp. 151--172, 1999.
  [Online]. Available: \url{http://dx.doi.org/10.1029/1998RG900006}
\BIBentrySTDinterwordspacing

\bibitem{loukas2015graph}
A.~Loukas, M.~Cattani, M.~Zuniga, and J.~Gao, ``Graph scale-space theory for
  distributed peak and pit identification,'' in \emph{IPSN}.\hskip 1em plus
  0.5em minus 0.4em\relax ACM/IEEE, 2015.

\bibitem{shuman2011chebyshev}
D.~I. Shuman, P.~Vandergheynst, and P.~Frossard, ``Chebyshev polynomial
  approximation for distributed signal processing,'' in \emph{DCOSS}.\hskip 1em
  plus 0.5em minus 0.4em\relax IEEE, 2011, pp. 1--8.

\bibitem{perraudin2016global}
N.~Perraudin, B.~Ricaud, D.~Shuman, and P.~Vandergheynst, ``{Global and Local
  Uncertainty Principles for Signals on Graphs},'' \emph{arXiv preprint
  arXiv:1603.03030}, 2016.

\bibitem{christensen2016introduction}
O.~Christensen, \emph{An introduction to frames and Riesz bases (Second
  Edition)}.\hskip 1em plus 0.5em minus 0.4em\relax Birkh{\"a}user, 2016.

\bibitem{grochenig2013foundations}
K.~Gr{\"{o}}chenig, \emph{{Foundations of Time-Frequency Analysis}}, ser.
  Applied and Numerical Harmonic Analysis.\hskip 1em plus 0.5em minus
  0.4em\relax Boston, MA: Birkh{\"{a}}user Boston, 2001.

\bibitem{shuman2012windowed}
D.~I. Shuman, B.~Ricaud, and P.~Vandergheynst, ``A windowed graph fourier
  transform,'' in \emph{Statistical Signal Processing Workshop (SSP), 2012
  IEEE}.\hskip 1em plus 0.5em minus 0.4em\relax Ieee, 2012, pp. 133--136.

\bibitem{shuman2015spectrum}
D.~I. Shuman, C.~Wiesmeyr, N.~Holighaus, and P.~Vandergheynst,
  ``Spectrum-adapted tight graph wavelet and vertex-frequency frames,''
  \emph{IEEE Trans. Signal Process.}, no.~16, pp. 4223--4235, 2015.

\bibitem{kovacevic2007life1}
J.~Kovacevic and A.~Chebira, ``Life beyond bases: The advent of frames (part
  i),'' \emph{IEEE Signal Process. Mag.}, no.~4, pp. 86--104, July 2007.

\bibitem{perraudin2014gspbox}
N.~{Perraudin}, J.~{Paratte}, D.~{Shuman}, V.~{Kalofolias}, P.~{Vandergheynst},
  and D.~K. {Hammond}, ``{GSPBOX: A toolbox for signal processing on graphs},''
  \emph{ArXiv e-prints}, Aug. 2014.

\bibitem{perraudin2014unlocbox}
N.~Perraudin, D.~Shuman, G.~Puy, and P.~Vandergheynst, ``{UNLocBoX A matlab
  convex optimization toolbox using proximal splitting methods},'' \emph{ArXiv
  e-prints}, feb 2014.

\bibitem{prusa2014large}
\BIBentryALTinterwordspacing
Z.~Pr{\r{u}}{\v{s}}a, P.~L. S{\o}ndergaard, N.~Holighaus, C.~Wiesmeyr, and
  P.~Balazs, \emph{The Large Time-Frequency Analysis Toolbox 2.0}.\hskip 1em
  plus 0.5em minus 0.4em\relax Cham: Springer International Publishing, 2014,
  pp. 419--442. [Online]. Available:
  \url{http://dx.doi.org/10.1007/978-3-319-12976-1_25}
\BIBentrySTDinterwordspacing

\bibitem{buades2005nonlocal}
A.~Buades, B.~Coll, and J.~M. Morel, ``A non-local algorithm for image
  denoising,'' in \emph{in CVPR}, June 2005, pp. 60--65.

\bibitem{chan2011augmented}
S.~H. Chan, R.~Khoshabeh, K.~B. Gibson, P.~E. Gill, and T.~Q. Nguyen, ``An
  augmented lagrangian method for total variation video restoration,''
  \emph{IEEE Trans. Image Process.}, vol.~20, no.~11, pp. 3097--3111, 2011.

\bibitem{lowrie2007fundamentals}
W.~Lowrie, \emph{{Fundamentals of Geophysics}}, 2nd~ed.\hskip 1em plus 0.5em
  minus 0.4em\relax Cambridge University Press, 2007.

\bibitem{durran2013numerical}
D.~R. Durran, \emph{{Numerical methods for wave equations in geophysical fluid
  dynamics}}.\hskip 1em plus 0.5em minus 0.4em\relax Springer Science {\&}
  Business Media, 2013.

\bibitem{strang1999discrete}
G.~Strang, ``{The discrete cosine transform},'' \emph{SIAM review}, no.~1, pp.
  135--147, 1999.

\end{thebibliography}


\end{document}